**Title**: Melding LLM and temporal logic for reliable human-swarm collaboration in complex environments

**Authors:**

Junfeng Chen[1], Yuxiao Zhu[2], An Zhuo[1], Xintong Zhang[2], Shuo Zhang[1], Guanghui Wen[3], Xiwang Dong[4], Meng Guo[1]*, Zhongkui Li[1]*

**Affiliations:**

[1]School of Advanced Manufacturing and Robotics, Peking University, Beijing & 100871, China.
[2]Division of Natural and Applied Sciences, Duke Kunshan University, Suzhou & 215316, China.
[3]School of Automaton, Southeast University, Nanjing & 210096, China.
[4]School of Automation Science and Electrical Engineering, Beihang University, Beijing & 100191, China.

*Corresponding author. Email: meng.guo@pku.edu.cn, zhongkli@pku.edu.cn.

**Abstract:** Robot swarms promise scalable assistance in complex and hazardous environments. Task planning lies at the core of human–swarm collaboration, translating the operator's intent into coordinated swarm actions and helping determine when validation or intervention is required during execution. In long-horizon missions under dynamic scenarios, however, reliable task planning becomes difficult to maintain: emerging events and changing conditions demand continual adaptation, and sustained operator oversight imposes substantial cognitive burden. Existing LLM-based planning tools can support plan generation, yet they remain susceptible to invalid task orderings and infeasible robot actions, resulting in frequent manual adjustment. Here we introduce a neuro-symbolic framework for long-horizon human-swarm collaboration that tightly melds verifiable task planning with context-grounded LLM reasoning. We formalize mission goals and operational rules as temporal logic formulas and admissible task orderings as task automata. Conditioned on these formal constraints and live perceptual context, LLMs generate executable subtask sequences that satisfy mission rules and remain grounded in the current scene. An uncertainty-aware scheduler then assigns subtasks across the heterogeneous swarm to maximize parallelisms while remaining resilient to disruptions. An event-triggered interaction protocol further limits operator involvement to sparse, high-level confirmation and guidance. In large-scale simulations with more than 40 robots executing 41 tasks with 155 subtasks in 11-minute missions, our system improves task success rates by 26% and increases completed tasks by 132% relative to state-of-the-art baselines. At the same time, it reduces operator interventions by 77% and lowers physiological stress by 49%. Deployment on a heterogeneous robotic fleet yields similar results while remaining robust to hardware-specific actuation and communication uncertainties. Together, these results support a formal and scalable paradigm for reliable and low-overhead human–swarm collaboration in dynamic environments.

## INTRODUCTION

Robot swarms can improve mission efficiency in complex environments by distributing sensing and action across many platforms (*1*). Parallel operation enables broad spatial coverage, rapid response to evolving conditions, and robustness to individual robot failures. In safety-critical domains such as search-and-rescue and firefighting, aerial vehicles provide situational awareness, while ground and legged robots can perform close-range inspection and hazard mitigation in areas unsafe or inaccessible for personnel (*2*–*6*). Realizing this potential, however, depends critically on reliable task planning for human–swarm collaboration. Task planning translates operator intent into coordinated swarm behavior while also defining when human confirmation or corrective intervention is required during execution. This role becomes especially challenging in long-horizon missions under dynamic scenarios, where unfolding events, shifting priorities, and environmental changes can invalidate earlier decisions and require continual replanning. At the same time, sustained operator supervision imposes substantial cognitive demands that scale poorly with mission duration and complexity. Existing methods are often tailored to narrow coordination primitives or specific missions, limiting their applicability in dynamic and unstructured environments (*7*, *8*). These limitations motivate human–swarm collaboration frameworks that combine reliable online task planning with timely and low-overhead human oversight

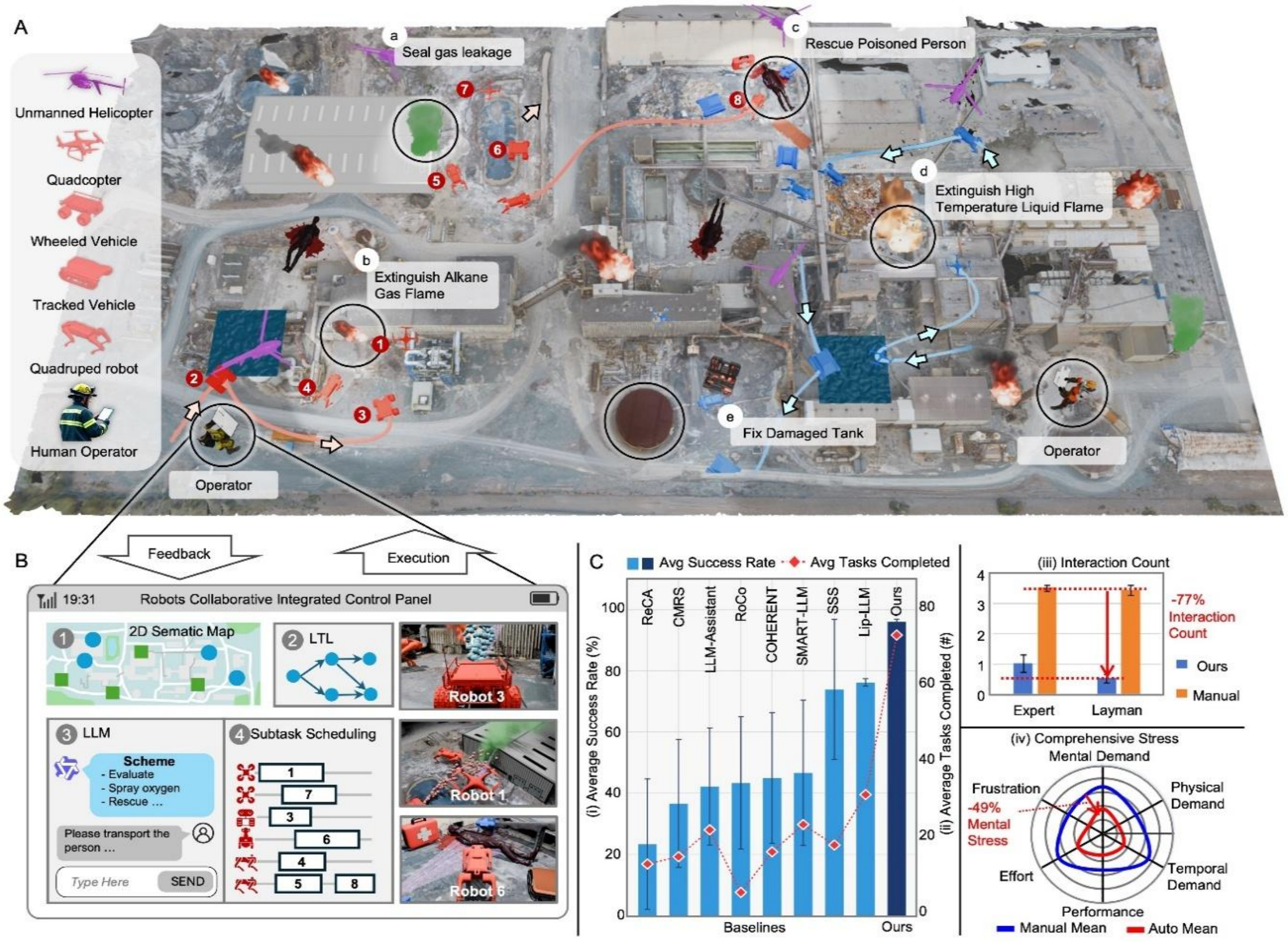

**Fig.1. An illustration of the proposed method for task planning of swarms. (A)** Operation scenario of 2 human operators and 17 robots (7 ariel, 5 legged, 5 tracked) during response missions. Exemplary tasks shown in (a)-(f) are discovered and handled online. Two distinct colored robot groups, each supervised by an operator, collaboratively perform extended sequential tasks, like extinguishing fire and then rescuing the injured personnels. **(B)** Human–swarm coordination: The integrated control panel provides real-time feedback and execution monitoring through multimodal perception, temporal-logic task comprehension, LLM-based reasoning for subtask generation and dynamic subtask assignment, facilitated by intuitive human supervision. **(C)** Performance evaluation: (i) Comparative analysis with baseline methods across task success rate. (ii) Number of finished tasks. (iii) Number of Human–swarm interaction. (iv) Psychological metrics with and without the proposed system.

Task planning for robot swarms is difficult because realistic missions require task interpretation, long-horizon reasoning, and combinatorial assignment. In representative deployments (shown in Fig. 1A), task specifications include rich temporal structure, such as event-triggered responses, safety rules, and operator directives. These requirements extend well beyond simple sequential or parallel compositions, thus deriving correct temporal ordering and dependencies is nontrivial (*9*). The difficulty further increases over long horizons, where small reasoning errors can propagate across successive decisions: generated subtask sequences may invoke unavailable actions, violate physical reconditions, or impose invalid temporal dependencies, leading to brittle execution (*10*). Even with a correct subtask set, allocating these subtasks across heterogeneous robots introduces a separate computational challenge. Subtask assignment is NP-hard and further constrained by synchronization requirements, resource limits, and heterogeneous capabilities (*11*). To reduce manual task modeling, recent approaches leverage LLMs to generate plans directly from natural-language instructions (*12–14*) and to translate multimodal requests into structured representations such as PDDL (*15*, *16*) and behavior trees (*17*). Hybrid pipelines that couple LLM outputs with classical planners further aim to impose structure and feasibility (*18–22*). However, such methods often degrade as the swarm size and task horizon grow, yielding incorrect task instances, flawed temporal dependencies or schedules that conflict with operational constraints and operator intent (*23*). In contrast, symbolic planning, especially temporal-logic-based methods, provides formal specification and verification (*24*), and supports explainable coordination pipelines for swarms (*25–27*), from formalizing operator intent (*28*) to enforcing temporal and resource constraints during assignment (*29*, *30*). Yet purely symbolic pipelines typically require substantial upfront modeling of tasks and rules, limiting adaptability in dynamic and partially observed environments. This tension motivates neuro-symbolic synthesis that melds temporal-logic guarantees with the contextual flexibility of LLMs (*31*, *32*), but reliable and scalable coordination over long horizons remains unresolved.

Beyond automated task planning, effective human–swarm collaboration under uncertainty remains a central bottleneck, particularly in determining when and how operators should validate plans and intervene during execution. In practice, no general-purpose criterion reliably predicts whether an autonomous plan will remain feasible and safe throughout execution. Human expertise therefore remains essential for validating constraint satisfaction interpreting evolving context, and assessing operational risk (*33*, *34*). However, most existing human–swarm interfaces and control paradigms do not explicitly support this validation role (*35*). Operators are often confined to static visualizations or post hoc plan review, rather than being integrated throughout the planning–execution loop for progress monitoring, timely intervention, and incorporation of real-time intent updates (*36*). These limitations are amplified in dynamic settings where information streams continuously from many robots, and often must be managed by multiple operators. As shown in Fig.1B, inputs also span heterogeneous modalities including voice, text, video, which calls for interaction protocols that are unified yet low-overhead. Without principled mechanisms for timely validation and intervention, some systems revert to teleoperation or manual replanning, forcing operators to redefine tasks, reason about dependencies, and reschedule execution (*37*). This creates severe cognitive bottlenecks; empirical evidence shows that supervising more than about three robots can substantially degrade operator performance, increasing sequencing errors and response latency (*38*). These limitations motivate low-overhead interaction protocols that elicit targeted human validation at the right moments. Such protocols should support sparse high-level intent and targeted oversight while preserving scalable autonomy in long-horizon and partially observed missions (*39*).

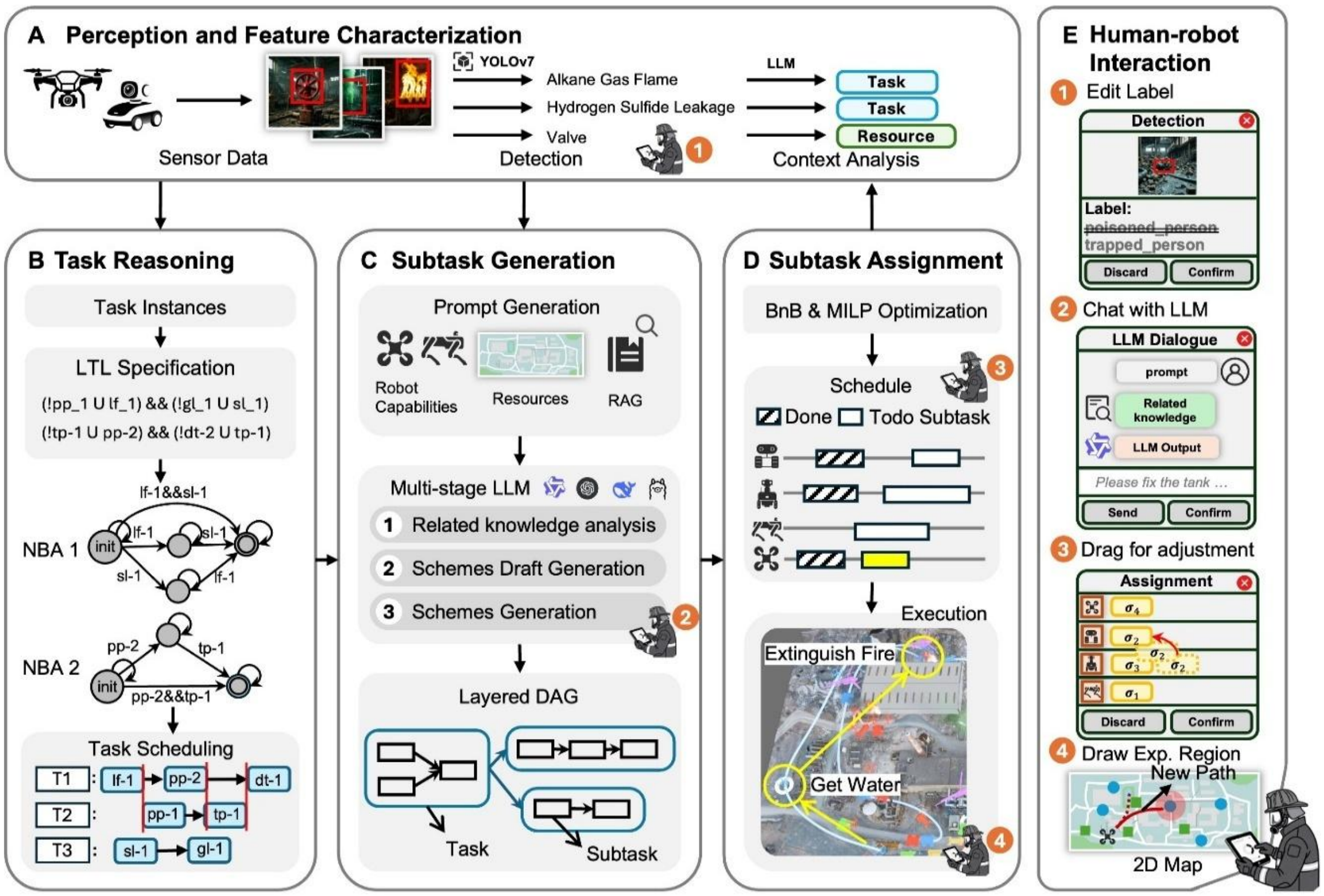

**Fig.2. Framework Overview. (A)** Environmental Perception: The swarm uses cameras and an LLM-based semantic characterization module to classify and interpret environmental features. **(B)** Task Reasoning: The task reasoning processes extracted task instances and applies automaton-based analysis to derive temporal constraints, generating a comprehensive task allocation scheme. **(C)** Subtask Generation: A multi-stage LLM produces a feasible subtask sequence that satisfies the identified temporal constraints, which can be executed by the robots. **(D)** Subtask Assignment: The subtask assignment selects the optimal subtask sequence for each task, assigns each subtask to a cooperating set of robots, and triggers predefined action primitives for execution. **(E)** Interactive Interface: An interface integrates human input and expertise during runtime, enabling dynamic adjustments to improve planning reliability and task execution.

To address these coupled gaps in long-horizon task planning and online human validation under uncertainty, we propose a neuro–symbolic framework that integrates formal task reasoning, context-grounded LLM decomposition, optimization-based assignment, and low-overhead human oversight, as shown in Fig. 2. Mission objectives and operational rules are encoded in Linear Temporal Logic (LTL), from which we construct a task automaton that captures admissible temporal orderings and logical dependencies. Conditioned on these formal constraints and environmental context, a multi-stage retrieval-augmented LLM translates high-level goals into grounded subtask sequences that satisfy mission rules while remaining adaptable to evolving observations. To maintain robustness in dynamic environments, we introduce a rolling-horizon optimization scheme that updates allocations and schedules as new tasks and observations arrive. We further develop an event-triggered interaction protocol that supports progress monitoring and mid-course plan validation while limiting operator input to sparse, high-level adjustments when needed. Experiments show that LTL achieves 100% accuracy of temporal planning, and the multi-stage LLM with RAG improves contextual task reasoning by 75% and reducing infeasible plans by 86%. In a large-scale 11-minute experiment, the system managed 41 tasks, handled 155 subtasks, and incorporated 140 human revisions. As shown in Fig. 1C, our system increases task success rate by 26% and the number of completed tasks by 132% relative to state-of-the-art baselines. It also reduces manual interventions by 77% and lowers physiological stress by 49%, indicating gains in autonomy and reductions in operator burden. The description details of the overall framework can be found in Movie S1.

## RESULTS

To evaluate the proposed framework in complex, partially observable environments, we performed four complementary studies. We first used large-scale simulations in a chemical-plant scenario to assess planning reliability and operational efficiency relative to representative baselines. We next evaluated human–swarm collaboration, testing whether the interface reduced operator workload while preserving effective human guidance under uncertainty. We then deployed the complete system in a realistic field environment (Fig. 5A(i)) to examine real-world feasibility with a heterogeneous robotic team and operators in the loop. Finally, we performed ablation studies to quantify the effects of different LLM backbones and the contribution of each core module to overall system performance.

### Planning reliability and efficiency

To evaluate planning reliability and efficiency, we tested the proposed framework in three simulated chemical-plant search-and-rescue scenarios with increasing difficulty, i.e., easy, medium, and hard. We created three scenarios of increasing difficulty by varying both the number and spatial distribution of environment features, i.e., the easy scenario contained 100 sparsely distributed features, the medium scenario included 200 evenly distributed features, and the hard scenario maintained 400 features clustered into several regions, yielding progressively higher feature-detection rates. Each difficulty-level experiment was evaluated 10 trials. All tasks in the environment were grouped into four categories, including fire suppression, casualty rescue, equipment maintenance, and leak containment, as detailed in the "Additional Implementation Details" section of the Supplementary Results. The swarm comprised 40 robots organized into five categories, including 4 unmanned helicopters (UHelis), 10 unmanned aerial vehicles (UAVs), 6 legged robots, 10 unmanned tracked ground vehicles (TUGVs), and 10 wheeled ground vehicles (UGVs). Each platform provided distinct sensing and actuation capabilities, as detailed in "Robot Behaviors and Motion Planning" in the Supplementary Hardware Experiment. A single operator supervised all human–robot interaction interfaces for five groups.

Over the full 11-minute simulation, the proposed framework successfully handled the mission at scale (Fig. 3E). In total, it decomposed 41 tasks into 155 cooperative subtasks across the full scenario. The four task categories—fire suppression, casualty rescue, equipment maintenance, and leak containment—appeared 11, 16, 5, and 5 times, respectively, together with 4 additional exploration tasks. All inter-task and inter-subtask temporal constraints were satisfied, as shown by the task- and subtask-level Gantt charts in Fig. 3E(i) and (ii). These results indicated that the framework maintained temporally consistent execution throughout a long-horizon mission while coordinating heterogeneous robots over many concurrent activities.

Then, it was worth noting that a conventional MILP formulation was not directly applicable to this setting because it presupposed a fully specified task graph and therefore could not infer tasks or subtasks from a high-level mission description. Even when a fixed subtask sequence was manually specified, the resulting joint allocation-and-scheduling problem remained combinatorially expensive. For example, in a simple mission with only 3 tasks, each decomposed into 5 subtasks, assigning the resulting 15 subtasks to 5 robots already yield a very large number of candidate assignment patterns before accounting for precedence, synchronization, and motion-feasibility constraints. In our tests, a commercial solver such as Gurobi (60) required more than 600 s for such instances, and computation quickly became impractical as problem scale increased. In contrast, the proposed framework successfully handled the full mission, demonstrating that the combined decomposition-and-assignment pipeline remained tractable at the scale of the full scenario.

To further illustrate dynamic task execution and the system's online adaptation capabilities (see Movie S2), we selected four representative events highlighting different task execution patterns. First, Fig. 3A showed a case in which the system successfully detected a fire and executed a task based on available resources. During a nominal fire-suppression episode, the system detected a fire (Fig. 3A(i)) at 25 s via the UAV onboard feature characterization module, with a latency of no more than 10 ms, and immediately uploaded the information to the operator's tablet. Subtask generation then produced two alternative subtask sequences within 12 s, and subtask assignment selected the optimal one within 40 ms, as shown in Fig. 3A(ii). Fig. 3B showed a case in which insufficient resource exploration prevented execution of the initial subtask sequence. For a high-temperature gas fire, the optimal strategy initially relied on a metal mesh, with water and soil as backups. After the UAV searched for 2 min without locating the mesh, the system triggered online adaptation and switched to the water-based plan. The UAV found a water source within 35 s, and the UGV completed retrieval and spraying in 30 s. Fig. 3C showed a case in which discovery of a new resource triggered adaptation of the original subtask sequence. For an electronic fire with high temperature, the initial plan used soil mounds, but during execution the UAV discovered a metal mesh and synchronized this information to the cloud, enabling the group-level LLM to generate a revised plan and complete the task successfully. Fig. 3D showed a case in which a more accessible resource was discovered during execution. A TUGV was originally assigned to retrieve water from location A, but the UAV later detected another source at location B. Re-running subtask assignment showed that using location B reduced the completion time from 40 s to 23 s, and the TUGV was redirected accordingly. In total, 37 such events were handled during the simulation, including 20 resource-exploration events, 6 inadequate-resource-exploration events, 5 new-resource-discovery events, and 6 closer-resource-utilization events, with no unresolved failures. More details were provided in the Supplementary Results and Movie S6.

The execution characteristics of the proposed framework for a representative robot group were summarized in Fig.3F. The trajectories in Fig. 3F(i) showed that the group coordinated to complete multiple tasks that emerged online. As shown in Fig. 3F(ii), the task reasoning, subtask generation, and subtask assignment components were invoked 10, 19, and 43 times, respectively. This increasing trend was consistent with the hierarchical design, i.e., task reasoning operated at a low frequency to interpret high-level goals and form mission-level task sequences, whereas subtask generation was triggered more often to decompose each task into executable subtasks, and subtask assignment was invoked most frequently because it performed real-time, rolling allocation at the lowest level, including within our online adaptation mechanism. Task reasoning required on average 3 s, reflecting that in dynamic settings missions were formed from roughly 7 tasks on average and could be reasoned about concurrently across groups. Subtask generation required 18 s on average; although it involved operator-in-the-loop interaction with the LLMs, the model runtime was reduced by our multi-stage reasoning procedure compared with deep thinking mode (20 s vs. 45 s). Subtask assignment had the fastest runtime, with an average of 0.6 s. This was due to the use of a receding-horizon allocation scheme, which typically batched around 4 tasks per update. Each task generated approximately 4 subtasks, leading to about 16 subtasks allocated per cycle. Overall, the mean reaction time per task was 30 s, enabling real-time responsiveness in the dynamic environment. The average execution times for the four task categories were 45 s for fire suppression, 40 s for casualty rescue, 38 s for equipment maintenance, and 35 s for leak containment, with the tasks related to fire suppression longer due to their longer subtask sequences. Finally, Fig. 3F(iii) summarized the frequency of human–swarm interactions, totaling 140 interface operations. Consistent with the interaction types in Fig. 2E, the operator most frequently performed perception-related inputs such as feature identification and verification followed by interactions that supported subtask generation, whereas exploration-related interventions were less common. Operator involvement was most pronounced during the second and third representative events at 4 and 5, to select viable alternatives and to guide exploration toward regions more likely to reveal missing resources.

We next compared the proposed framework with representative baseline methods spanning both centralized and distributed architectures. Then Within centralized approaches, we evaluated end-to-end methods in which the LLM directly generated executable action sequences for each robot. These included CMRS (*40*) and LLM-Assistant (*41*), the latter of which incorporated a human verification. We also tested centralized hierarchical methods where each planning layer, namely task reasoning , subtask generation, and subtask assignment, was implemented with a separate LLM module. Representatives in this category were COHERENT (*19*) and ReCA (*21*), with ReCA designed to include human-in-the-loop oversight. A distinct centralized strategy employed hybrid model-based and LLM-based components. Lip-LLM (*42*), for instance, used a Directed Acyclic Graph (DAG) for temporal planning, an LLM for task reasoning, and a formal optimization model for scheduling. We also examined distributed paradigms typical for swarms. For instance, ROCO (*20*) adopted a distributed end-to-end LLM approach, where each robot used its own LLM for local task planning and then iteratively exchanged proposals with neighbors to reach a global consensus. In contrast, the SSS framework (*22*) employed a distributed hybrid architecture, i.e., each robot first computed a local plan using a model-based temporal planner and an LLM-based reasoner, akin to Lip-LLM, before engaging in a distributed dialogue protocol to converge on a consistent global solution. To evaluate the performance of these methods, we measured several key metrics, i.e., the success rate as the ratio of completed tasks to the total tasks; the efficiency as the number of tasks completed per unit time; the causes of planning failures, categorized into incorrect temporal ordering, incorrect subtask sequences, or subtask assignment failures; the planning consistency evaluated by whether repeated trials under identical conditions yield identical plans; and the planning quality measured by the discrepancy between the intermediate outputs and the ground truth from the experts.

Quantitative results were summarized in Fig. 1C. Unlike the simulation study, we increased the number of tasks in the comparison experiment to provide a more stringent evaluation of the proposed framework. Our method achieved the highest average success rate with the smallest standard deviation (96±0.9)%, while end-to-end LLM-based methods performed below 26% success rate with higher variance compared to the best performance of other baselines (76±1.2)%, reflecting the robustness in different scenarios (see Fig.1C(i)). Our method completed the most tasks at 72.6, followed by the hybrid LiP-LLM with 31.3, supporting the advantage of combining model-based structure with LLM-based reasoning in complex settings. Furthermore, we evaluated the planning efficiency, which showed a 132% improvement in our method compared with the best performance of other baselines as illustrated in Fig. 1C(ii), highlighting its high planning efficiency. As shown in Fig. 3G(i), our method failed only once during the stage of subtask generation. In contrast, end-to-end LLM-based methods failed frequently across all categories (average on 17.6, 16.1 and 6.2), primarily in subtask generation (average on 40% of total errors) and subtask assignment (average on 44 % of total errors), which was because LLMs were prone to hallucinations in long-horizon task reasoning and were ill-suited for solving combinatorial optimization problems. The hybrid LiP-LLM, using an optimization-based scheduler, showed fewer assignment errors (1 vs . 5, 5, 8, 4, 7, 6, 6) but was still outperformed by our method in task reasoning and subtask assignment, underscoring the advantage of our integrated LTL-based grounding and multi-stage LLM reasoning framework. Fig. 3G(ii) showed that our method and LiP-LLM achieved the highest planning consistency, while purely LLM-based methods exhibited lower stability (8 out of 10). Finally, Fig. 3G(iii) confirmed that our method maintained highest planning quality consistently across all difficulty levels. This robustness resulted from our guided, multi-stage reasoning process and optimization-based subtask assignments, which together ensured deterministic and reliable outputs. Additional reliability metrics further supported this conclusion, i.e., incorporating the exploration mechanism increased the overall task success rate by 28.5%, and our method achieved the longest mean time to failure (610 s) among all baselines, indicating reduced susceptibility to failures. Moreover, the resulting plan lengths closely matched the ground-truth trajectories with a difference in the length less than 0.26, suggesting that the synthesized subtask sequences were well calibrated rather than conservative or verbose. We also observed the highest planning predictability of nearly 98%, meaning the proposed subtask plans were easier for an operator to anticipate and assess, further indicating that the generated plans were coherent and reasonable. Additional details were provided in the Supplementary Results.

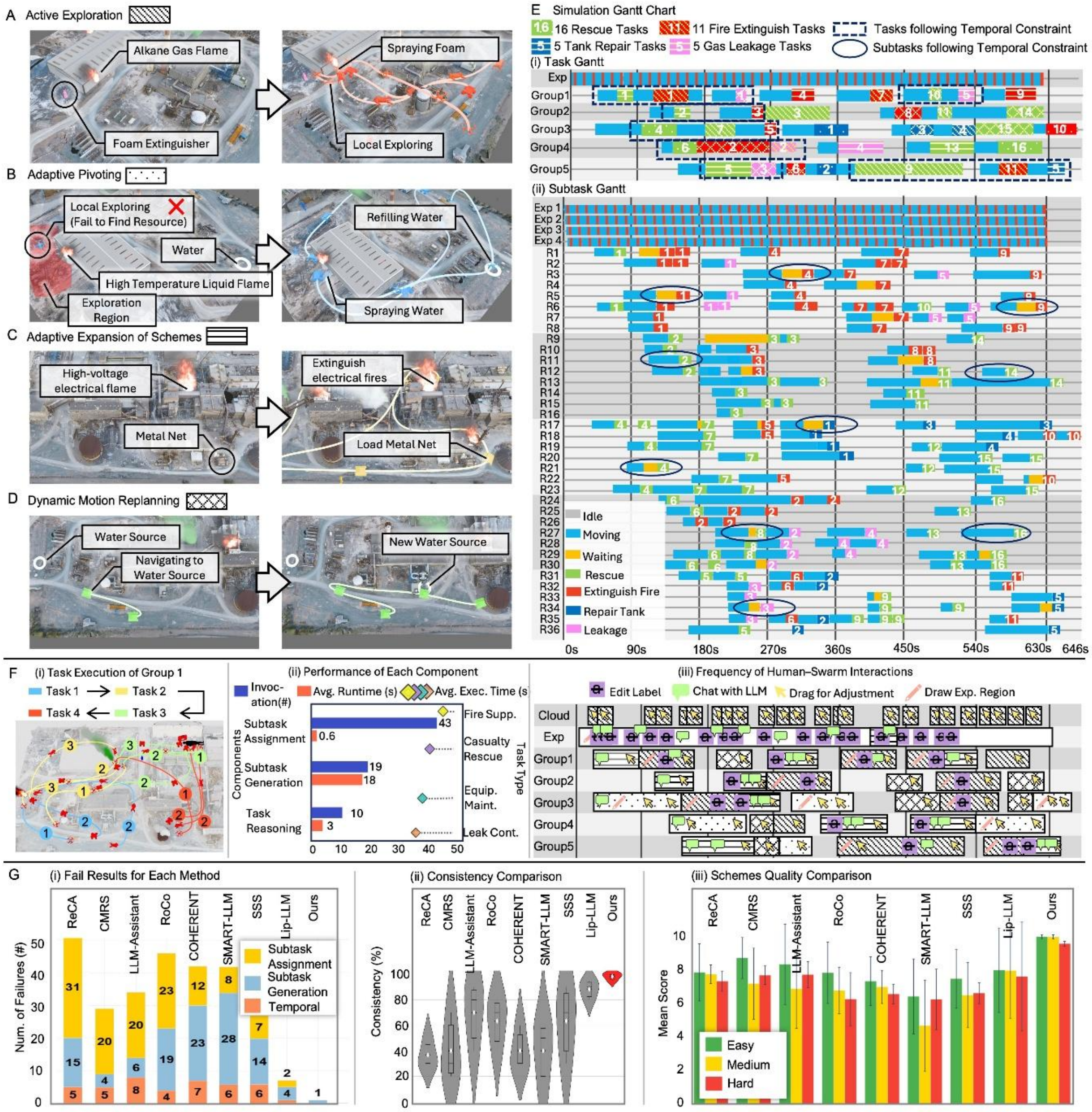

**Fig.3. Planning reliability comparison. (A)** Event 1: The resource required for the current foam extinguishing subtask is successfully found and acquired through exploration. **(B)** Event 2: The initially planned fire extinguisher for the subtask is not found during exploration, prompting a dynamic switch to an alternative water-fetching subtask. **(C)** Event 3: A metal net is discovered during task execution, leading to the adoption of this more efficient method for fire suppression. **(D)** Event 4: During a water-fetching subtask, a closer water source is found, and the plan is optimized to use this proximal resource. **(E)** Overall process: (i) Task execution states for each group, including four task types and their temporal relationships. (ii) Subtask execution states for each robot, detailing the subtask sequence and the temporal relationships between subtasks. **(F)** Overview of system performance: (i) Task execution details for a specific group, with distinct trajectory colors representing different completed tasks. Numbered markers along each trajectory indicate the execution order within a task, where identical indices correspond to parallel actions and smaller numbers represent higher execution priority. (ii) Performance of the three main components, including the number of invocations per component, average execution time per invocation, and the number for each of the four task types. (iii) Overview of event occurrences, along with the human-robot interactions employed for each event. **(G)** Reliability metrics: (i) Failure cause analysis for all method. (ii) Plan consistency across 20 runs. (iii) Planning quality is evaluated across difficulty levels, where higher scores indicate better consistency.

## Human Robot Collaboration

To evaluate whether the proposed framework could reduce operator workload while preserving the contribution of human intent to task planning, we conducted a human-swarm interaction experiment. As shown in Fig. 4A(i), four interactive interfaces were integrated throughout the software architecture to guide the planning process. These interfaces enabled the manual confirmation in the text input for the task reasoning, the drag-and-drop task selection for the subtask assignment, the manual region specification for exploration tasks and the teleoperation for task execution. Then we recruited 18 volunteers for the user study, illustrated in Fig. 4A(ii), among whom 3 had relevant professional expertise, while the remaining 15 were non-specialists. All participants completed three scenarios of increasing difficulty after receiving standardized training provided to ensure familiarity with the system. During the experiment, we continuously monitored several physiological indicators, including heart rate (HR) and galvanic skin response (GSR) to assess the stress levels, as depicted in Fig. 4A(ii). After each trial, participants were asked to complete the NASA Task Load Index (NASA-TLX) (*43*) to quantify perceived workload and the System Usability Scale (SUS) (*44*) to provide subjective feedback on the usability. As illustrated in Fig. 4B(i), each questionnaire comprised approximately 20 items answered; the resulting scores were aggregated and analyzed offline. Fig. 4B(ii) further displayed the temporal evolution of HR and GSR throughout the experiment. Notably, the critical situations, such as when resources were initially unavailable and operators needed to direct robots to explore and identify alternatives, were accompanied by concurrent increases in HR and GSR, reflecting elevated stress associated with managing unexpected contingencies. By contrast, tasks that could be executed immediately with minimal operator involvement elicited comparatively stable physiological responses. We also annotated the traces with representative events to highlight how task difficulty and urgency produced distinct stress profiles. Beyond human-in-the-loop verification, our human–swarm collaboration framework also provided an automaton-visualization interface for monitoring task execution. The full questionnaire forms were provided in the Supplementary Materials.

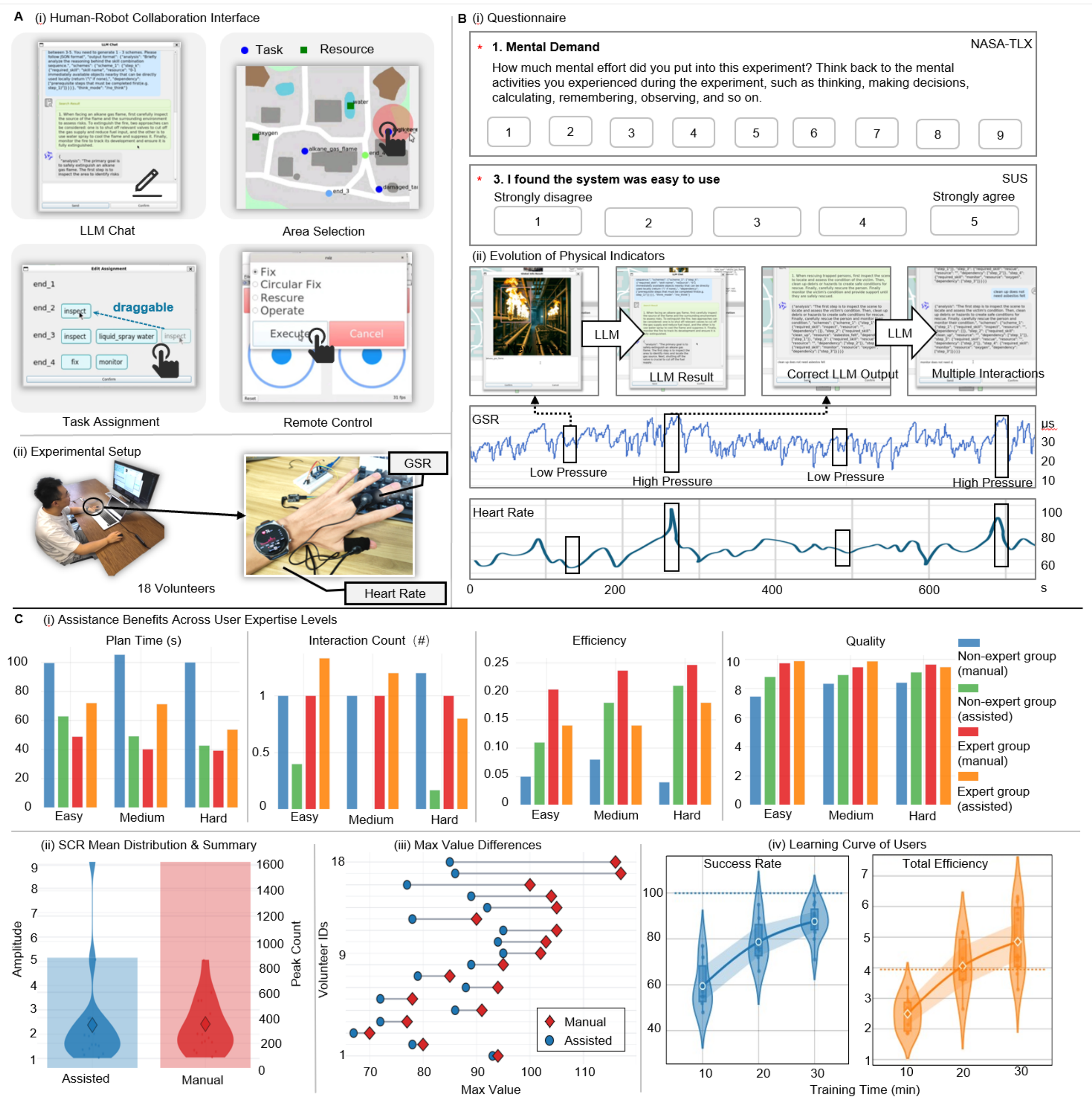

**Fig. 4. Human-swarm collaboration experiment. (A)** Human–robot interaction and experimental setup of volunteer study: (i) Four kinds of human–swarm interaction interfaces. (ii) An experiment with 18 volunteers recording synchronous galvanic skin response (GSR) and heart rate (HR) via professional equipment. **(B)** Post-experiment: (i) NASA-Task Load Index (NASA-TLX) and System Usability Scale (SUS) questionnaires were asked to administer after the experiment. (ii) Curves of GSR and HR throughout the experiment, with corresponding events at different peaks. For instance, when subtask sequences can be successfully executed without human intervention, the pressure is lower; whereas, when subtask sequences fail during exploration and the operator interacts with the LLM again, higher pressure is experienced. **(C)** Human–robot collaborative performance: (i) Comparison between expert and non-expert groups under fully-manual operation and operation with our system. (ii) Average SCR for all volunteers \ under manual (red) and system-assisted (blue) conditions. (iii) Peak heart-rate statistics for all volunteers: maximum heart rate during manual operation (red) versus maximum heart rate when using our system (blue). (iv) Learning curve of users with increased training time.

As quantitatively summarized in Fig. 4C(i), the performance of the expert group under manual planning and system-assisted planning was compared, along with that of the non-expert group under manual and system-assisted conditions. The results indicated that our system improved planning quality (7.9 vs. 9.0), across all difficulty levels for both user groups. In terms of planning time, non-experts exhibited a substantial reduction from 101 s to 48 s when using our system. Planning efficiency, however, was enhanced on average 220% for all participants with system support. Notably, the interaction frequency with the system differed between groups, i.e., non-experts tended to engage less extensively (on average 0.3) due to their unfamiliarity with the domain, whereas experts leveraged the interactive features more (1.1 vs. 0.3), leading to more effective collaboration and thus greater performance in both planning quality (9.8 vs. 9.0). Furthermore, questionnaire results in Fig.4C(ii) indicated the reduced perceived workload with system support (on average 2.3 vs. 2.6). Moreover, Fig. 4C(iii) showed that all participants exhibited the lower HR when assisted by the system. These physiological findings, together with the recorded GSR demonstrated that our framework effectively lowered both cognitive workload by 11.5% and psychological stress by 11.2%. Fig. 4C(iv) showed that both success rate (59.4% vs. 78% vs. 87.6%) and efficiency (2.5 vs. 4.0 vs. 4.8) improved steadily via the proposed system with increasing training time across all participants. In addition, as presented in Fig. 1C(ii), the system reduced the average number of the human–swarm interactions by 77% from 3.49 to 0.79. Finally, Fig. 1C(iii) reported a consistently lower subjective workload than manual planning in all dimensions, with an average reduction of 49% when using our system from NASA-TLX ratings. A detailed analysis of the role each module plays in our system, compared to fully manual planning, was provided in the Supplementary Results section, "Additional Results on Human–Swarm Interaction," and could also be seen in Movie S3. Finally, this interface enabled operators to track execution progress and to verify, in real time, adherence to the LTL-specified guarantees until all tasks reached their terminal states. Further details were provided in the Supplementary Results section “Additional Results on Execution Monitoring” and see Movie S5.

## Real-world Deployment

To validate the proposed system on hardware platforms, we constructed a miniature experimental site measuring 16 × 14 m, which modeled an industrial plant in Fig. 5A(i). The scene included industrial structures, pipelines, and multiple emergency tasks, such as 4 fire outbreaks, 3 injured personnel, and 1 chemical leak from storage tanks. To support task execution, we placed key resources like first-aid kits, fire hydrants, and extinguishers within the environment. Our robotic fleet consisted of 2 drones for aerial exploration, 4 UGVs and 2 leggeds which were managed by 3 operators, each equipped with a tablet. The LTL model was deployed on a cloud-hosted desktop workstation, while the LLM with Qwen3-8B ran on each edge tablet. We implemented a unified software and communication stack based on a three-layer cloud–edge–end architecture to command the heterogeneous swarm in complex environments (Fig. 5A(ii)). The cloud server coordinated cross-group tasks at 6 times and maintained the global task state, dispatching group-level task allocations. Each group was managed by an operator tablet as the edge node, which handled rapid on-site processing at 44 times. The end layer consisted of individual robots that autonomously executed assigned subtasks, using onboard action primitives and local motion planners. Communication across these platforms was facilitated via a WiFi-6 router–configured wireless local area network, and we employed a mixed-protocol design (ROS1, ROS2, and TCP/UDP-based bridging) to ensure the low communication latency (50 ms) and interoperable connectivity across the swarm. For further details, the implementation of the robot-level planners was provided in the “Hardware Experiment” section of the supplementary materials. Deployment specifics of each software component could be found in the “Software Architecture” section under Hardware Experiment, while the network configuration was detailed in the “Communication Network” subsection of the supplementary Hardware Experiment.

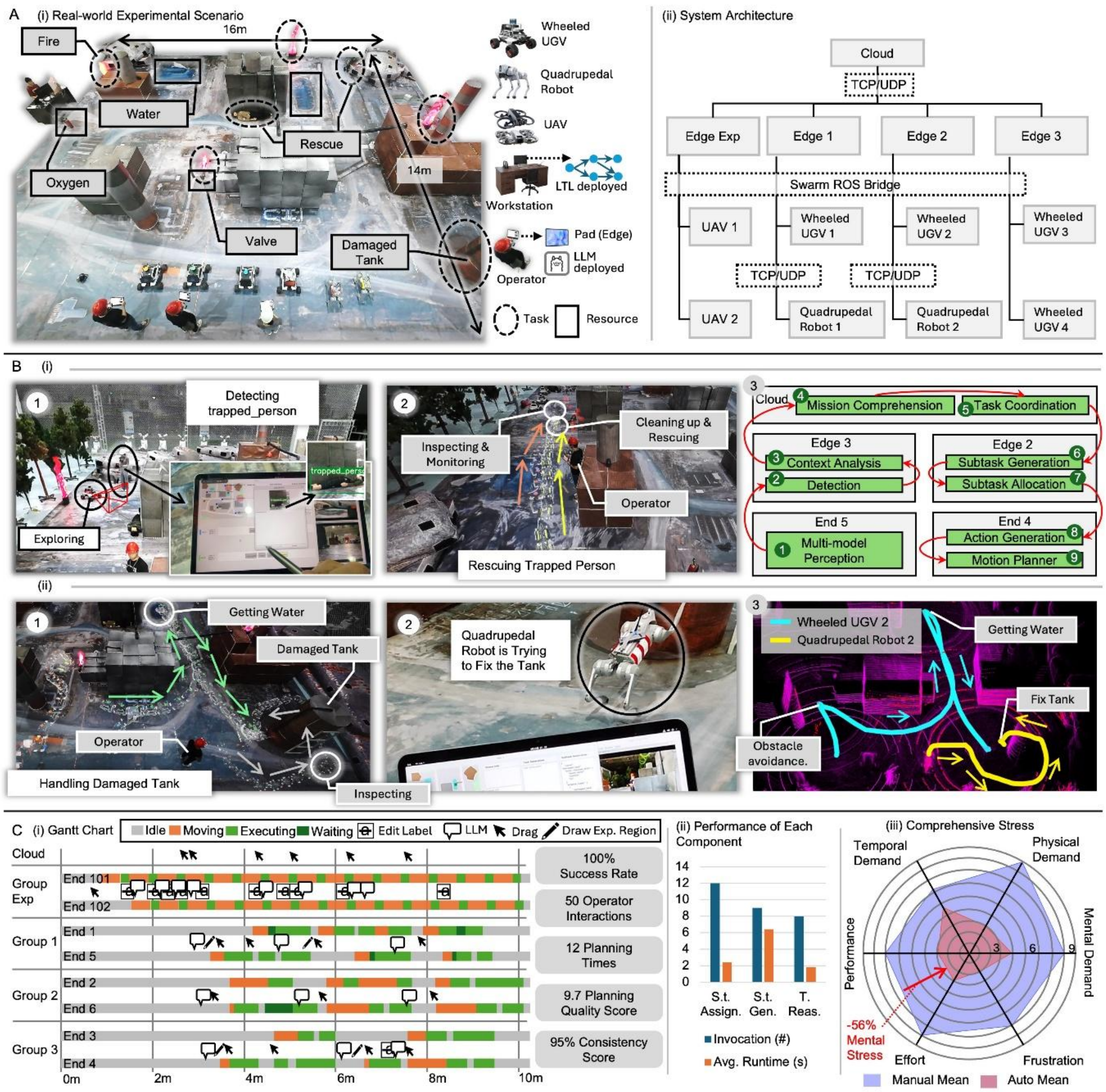

**Fig.5. Complex real-world experiment. (A)** Software and hardware setup: (i) In a 16m ×14 m miniature industrial plant, three operators, equipped with tablet interfaces, supervised a heterogeneous robot team comprising 2 UAVs, 4 UGVs, and 2 legged robots. (ii) LTL–based task reasoning runs on the cloud server, while LLM-based subtask generation and subtask assignment are deployed on each tablet. The task-planning components are deployed on a cloud–edge–end architecture, with inter-module communication facilitated by multiple network protocols. **(B)** Representative frames: (i) One group fails to complete a task, and collaborates with another group via cloud coordination. (ii) The operator remotely controls a legged robot to perform a dangerous task, with real-time video feedback. **(C)** Physical experiment performance: (i) Overview of each robot's system status, operator involvement, and performance summary. (ii) Invocation count and average execution time for each module in the system. (iii) Psychological metrics comparing performance with and without the proposed system in hardware experiments.

The full hardware trial lasted 10 min, during which the swarm completed 8 tasks comprising 33 subtasks. Across the site, 12 environmental features were detected and correctly classified. On average, each operator intervened 16 times through the tablet interface. The cloud-hosted LTL planner was executed 6 times to update mission-level priorities and allocations, while the on-device LLM was invoked 17 times to synthesize the plans. The trial began with the 2 UAVs performing continuous overwatch for 10 min, during which they discovered 8 features via the semantic perception pipeline: 2 major fires, 2 minor fires, 3 injured personnel, and 1 storage-tank leak requiring repair. Given these observations, the LTL specification enforced a safety-critical ordering, i.e., prioritizing casualty rescue first, followed by suppression of major fires, after which the minor fire and tank-repair tasks were executed in parallel. The cloud then dispatched the resulting task set to 3 groups, with each group handling 3 tasks on average. Within each group, the operator interacted with the LLM for an average of 14 s to generate candidate plans for each assigned task. On average, 3 candidate plans were generated for each task, and the selected plan contained 4 subtasks. These plans were then mapped to individual robots by the edge-level assignment module with average runtime of 0.5 s, yielding an average subtask execution time of 20 s and an average task completion time of 82.5 s.

To illustrate hardware execution in more detail, Fig. 5B further highlighted two representative episodes that illustrated cross-group coordination and human–robot collaboration. In Fig. 5B(i), at 420 s, a UGV executed a fire suppression task while simultaneously detecting an injured person. The robot captured an image with its onboard camera, which was processed through the semantic recognition pipeline via YOLO package. This information was transmitted to the edge device within 1 s. The feature characterization module in the edge node identified the task as a casualty rescue and notified that the UGV in the group lacked rescue capabilities. Consequently, the task was sent to the remote workstation within 2 s, where the task allocation reassigned it to another group, which was already engaged in the task of fire suppression. The task was then updated on the group's tablet, inserted after the ongoing fire suppression task. The subtask generation and assignment took 10 s to develop a subtask sequence, which included using the UGV to transport the injured person and the legged for the rescue. Upon completion of the fire suppression task at 490 s, the group’s UGV and legged executed the rescue task according to the generated plan. Fig. 5B(i)(3) illustrated the flow of information and task execution. In Fig. 5B(ii), another group concurrently addressed a storage-tank leak repair. Within 21 s, the group’s tablet generated a plan in which a legged performed the repair while a UGV monitored the repair status. At 220 s, the legged approached the leaking tank while streaming live video at 1 Hz to the operator, enabling real-time monitoring on the tablet. Using a dedicated legged interface (Fig. 5B(ii)(2)), the operator selected a pre-programmed “fix” primitive, prompting the robot to seal the breach using its foot with an end-to-end latency of under 20 ms. The resulting trajectory in Fig. 5B(ii)(3) demonstrated the framework’s ability to support safe human-swarm collaboration, where robots reliably executed low-level actions without collisions in hazardous environments. Full execution traces are available in Movie S4.

Fig. 5C summarized the quantitative outcomes of the hardware experiment. The proposed framework achieved a 100% success rate, and each trial required an average of 50 operator interactions. The average end-to-end planning time was 21.6 s based on 10 trials, while the resulting subtask sequence reached a planning quality score of 9.7 and a consistency of 95%. These results demonstrated that our system was capable of achieving high performance in real-world environments, which inherently introduced significant uncertainty, particularly in perception and motion execution, when compared to the simulation. This success could be attributed to two key factors. First, for perception, we improved the system’s sim-to-real capability by augmenting the YOLO network training with real-world images, in addition to simulation images, to enhance the model’s generality. Second, to address the real-world challenges of localization errors for motion planning, we employed a motion capture system in conjunction with the Fast-Lio localization package for sensor fusion to improve localization accuracy. Additionally, the use of the Move Base navigation stack ensured the reliable collision and obstacle avoidance, further enhancing the system's robustness in real-world environments.
These statistics followed the same trends observed in simulation, which suggested that the planning and execution pipeline transferred to the hardware without noticeable degradation. The three key components were triggered 6, 7 and 10 times, respectively, and their activation patterns matched those in simulation. To assess operator workload and usability in a realistic setting, all participant additionally conducted one full trial in a fully manual mode without the system assistance, and then completed NASA-TLX and SUS questionnaires for both conditions. Compared with fully manual operation, our system reduced perceived workload by 56% from 7.5 to 3.3. This reduction was more apparent than the workload reduction in simulation from 56% to 49%. Together, these results showed that the proposed framework preserved planning reliability on hardware while reducing operator cognitive burden when autonomy alone was insufficient, which supported its use for swarms in complex and real environments.

### Impact of Models Choice and Component Ablations

To evaluate the impact of choosing different LLM models under our proposed framework, we conducted five trials for each model under the above scenario with the dynamically-generated task, comparing models including DeepSeek, GPT-5, GPT-4o, Gemma, Llama, and Qwen3 via Ollama (*45*). As summarized in Table 1 upper, all models achieved 100% success rate in the module of context analysis except Qwen3-1.7b model. In the subtask generation, however, DeepSeek-671b operating in a deep-thinking mode, required the longest planning time of 18.63 s but achieved the highest planning quality of 9.2, suggesting that extended reasoning improved plan quality on complex tasks. Furthermore, scaling the parameter size within the Qwen3 family led to a clear trade-off, i.e., larger models increased the inference time but substantially improved planning quality and the number of tasks completed. Moreover, ablation studies in Table.1 bottom quantified the contribution of each component for our framework. The removal of RAG module had the most detrimental effect, resulting in the lowest planning quality and fewest completed tasks (5.4 and 47.3), underscoring its critical role in providing contextual knowledge. Omitting human verification also reduced the plan quality from 9.8 to 5.9, confirming the importance of human-in-the-loop refinement. The exclusion of the related knowledge analysis (MS-p1) or subtask guidance (MS-p2) modules led to comparable, discernible performance degradation (97.6 vs. 65.1 vs. 69.7 in terms of average finished tasks). Collectively, these results validated our framework, i.e., RAG provided essential background knowledge, the structured analysis modules supported accurate interpretation and reasoning, and human verification improved output robustness, all of which were critical for reliable operation in dynamic and unknown environments.

## DISCUSSION

This study shows that reliable long-horizon human–swarm collaboration can be improved by tightly coupling formal task structure, context-grounded language reasoning, and online human oversight. The proposed neuro-symbolic framework combines Linear Temporal Logic–based task specification, multi-stage LLM reasoning, and rolling-horizon optimization to support verifiable planning and adaptive execution in complex environments. Across both simulation and hardware experiments, this combination improved planning reliability, reduced operator burden, and maintained robust performance under dynamic conditions. Rather than relying on narrowly engineered coordination routines for specific scenarios, our results suggest a more generalizable paradigm for human–swarm task planning in uncertain and evolving environments.

**Table 1: Study on the impact of different LLMs and components of our method.** SR: Success Rate; PT: Planning Time (s); PQ: Planning Quality; Tasks: Number of Finished tasks. ↑ indicates the higher the better, and ↓indicates the lower the better

| Model | Context Analysis | | Task Reasoning | | | |
|---|---|---|---|---|---|---|
| | SR↑ (%) | PT↓ (s) | SR↑ (%) | PT↓ (s) | PQ↑ | Tasks↑ (#) |
| **Model Performance** | | | | | | |
| **Deepseek:671b** | 1.00 | 7.40 | 1.00 | 18.63 | 9.2 | 35 |
| **GPT-5** | 1.00 | 2.60 | 1.00 | 6.70 | 8.6 | 44 |
| **GPT-4o** | 1.00 | 2.63 | 1.00 | 6.30 | 6.8 | 46 |
| **Gemma3:12b** | 1.00 | 4.05 | 1.00 | 18.70 | 8.2 | 5 |
| **Llama3:1.8b** | 1.00 | 2.41 | 0.14 | 2.83 | 4.2 | 60 |
| **Qwen3:235b** | 1.00 | 7.22 | 1.00 | 9.83 | 6.7 | 35 |
| **Qwen3:14b** | 1.00 | 4.21 | 1.00 | 11.58 | 8.3 | 44 |
| **Qwen3:8b** | 1.00 | 1.31 | 0.85 | 7.16 | 6.2 | 30 |
| **Qwen3:1.7b** | 0.73 | 0.66 | 0.00 | 1.70 | 2.6 | 28 |
| **Ablation Study** | | | | | | |
| **Ours** | - | - | 1.00 | 15.20 | 9.8 | 97.6 |
| **w/o RAG** | - | - | 1.00 | 4.25 | 5.4 | 47.3 |
| **w/o MS-p1** | - | - | 1.00 | 2.72 | 7.1 | 65.1 |
| **w/o MS-p2** | - | - | 1.00 | 3.57 | 7.8 | 69.7 |
| **w/o Human** | - | - | 1.00 | 2.51 | 5.9 | 57.4 |

Furthermore, a key limitation of the present work is that it operates primarily at the strategic planning level and assumes reliable execution of abstract actions. In practice, this is optimistic, as realizing abstract actions remains challenging. Predefined meta-actions, such as acquire_resource (water), presume physical capabilities beyond current state-of-the-art, i.e., legged robots struggle to manipulate equipment while moving in unstructured terrain (*46*), and aerial vehicles face strict payload limits (*47*). This embodiment gap becomes especially apparent when abstract plans are instantiated as physical motions, where generated trajectories can exceed hardware limits, making execution fidelity dependent on each platform's physical intelligence. Future integration with embodied AI, particularly environmental affordance learning (*48*, *49*) and physics-informed skill synthesis (*50*), is essential to close this gap, transforming our coordination framework from a research prototype into a deployable solution for safety-critical applications such as chemical-plant emergency response.

Another important limitation stems from our assumption of continuous communication across the cloud-edge-end hierarchy. The current framework relies on perfect, real-time bidirectional links, overlooking practical constraints like bandwidth, latency, and adversarial jamming in contested environments. This is particularly critical in disaster scenarios where communications are often degraded or denied (*51*). To mitigate this, one promising direction is to integrate intermittent-communication strategies, such as scheduled rendezvous protocols (*52*). Cloud-edge synchronization would occur at pre-planned spatiotemporal waypoints, while edge-end coordination would follow similar scheduled meetings for local data exchange. Although introducing bounded latency, preliminary validation confirms that task integrity can be maintained through these predictable communication windows. Our modular architecture allows this enhancement without a fundamental redesign, enabling the framework to address such operational constraints as needed.

More broadly, the modular structure of the framework makes it well suited for adaptation to scenario-specific multi-robot systems. Rather than redesigning the full planning pipeline for each new application, developers can reuse the same coordination architecture while replacing domain-specific protocols, symbolic constraints, and meta-action libraries. For example, in forest-fire prevention or inspection missions, the framework could be adapted by redefining the relevant task rules, environmental resources, and executable action abstractions for that domain. This modularity may reduce the engineering effort required to transfer the system across application settings, while preserving the same overall logic of formal specification, grounded reasoning, adaptive allocation, and selective human oversight.

# METHODS

## System Architecture

As illustrated in Fig. 2, the proposed system implements a neuro-symbolic task planning framework for robot swarms. The architecture consists of the interconnected modules designed to ensure reliable and interpretable planning while actively incorporating human intent. First, the module of perception and feature characterization identifies environmental features as either task or resources, as displayed in Fig. 2A. Subsequently mission objectives from human operators are formally encoded into LTL specifications in the "Task Reasoning" section, which are then translated into a corresponding automaton structure for verifiable temporal sequencing (see Fig. 2B). These high-level tasks are subsequently decomposed into executable subtask sequences, which is explained in the "Subtask Generation" section and Fig. 2C. The resulting subtasks are then allocated to individual robots through the "Subtask Assignment" section and in Fig. 2D. Throughout this pipeline, a human-swarm collaboration interface enables operators to monitor the progress, validate intermediate outputs, and make necessary adjustments via multimodal inputs (see Fig. 2E). Supplementary Methods provide the details of the above components, the online adaptation mechanism, and hardware-related system aspects.

## Task Reasoning

### *Mission Specification and Reasoning via LTL*

At any given time step $t$, multiple missions may be active simultaneously, i.e., $\varphi_k \in \Phi_t$, where $\Phi_t \triangleq \{\varphi_1, \cdots, \varphi_K\}$ is the set of missions and $\mathcal{K} \triangleq \{1, \cdots, K\}$. Each mission consists of a set of tasks to be performed and the temporal dependency among the tasks. For every $\varphi_k$, the success of mission execution relies not only on nominal tasks but also on managing complex temporal dependencies among tasks. These constraints are formally captured by translating natural language requirements into LTL formulas, resulting in the specification:

$$\varphi_k \triangleq \varphi_{\text{norm}} \bigwedge_{\bar{e} \in \bar{E}} \square(\varphi_{\text{obs}}^{\bar{e}} \rightarrow \Diamond \varphi_{\text{rep}}^{\bar{e}}) \quad (1)$$

where $\varphi_{\text{norm}}$ is a syntactically co-safe LTL (sc-LTL) formula for the nominal tasks such as exploration by sequential visit, $\bar{E}$ represents the set of observable events, and for each event $\bar{e} \in \bar{E}$, $\varphi_{\text{obs}}^{\bar{e}}$ is a propositional formula that signals its occurrence and $\varphi_{\text{rep}}^{\bar{e}}$ is a sc-LTL formula that defines the desired response. This ensures that every observed event $\varphi_{\text{obs}}^{\bar{e}}$ eventually triggers its corresponding response $\varphi_{\text{rep}}^{\bar{e}}$. Both the nominal task and the response task can be equivalently transformed into a Non-deterministic Büchi Automaton (NBA) by (*54*) for each mission $\varphi_k$, i.e., $B_{\varphi_k} \triangleq (Q^k, Q_0^k, \Sigma^k, \delta^k, Q_F^k)$, where $Q^k$ is the set of states, $Q_0^k \subseteq Q^k$ is the set of initial states, $\Sigma^k \triangleq 2^{AP}$ over atomic propositions $AP$ is the input alphabet, and $Q_F^k \subseteq Q^k$ is the set of accepting state. A feasible run $\pi_k \triangleq q_0 \xrightarrow{\omega_1} q_1 \xrightarrow{\omega_2} \cdots \xrightarrow{\omega_n} q_n$ in the NBA $\mathcal{B}_{\varphi_k}$ is a sequence from an initial state $q_0 \in Q_0^k$ to an accepting state $q_n \in Q_F^k$, with each task $\omega_i \in \Sigma^k$ satisfying the transition relation $\delta^k$ and thus the LTL specification. Further examples of different mission specifications and the associated LTL formulas are included in section of "Simultaneous Task Decomposition and Group Assignment" of Supplementary Methods. Moreover, formal definition of LTL and sc-LTL are omitted (*53*) here and detailed in Supplementary Method.

***Group Assignment via Multiple automaton guided Search***

In the complex environments, coordinating the swarm under temporal task specifications remains challenging. Conventional approaches often rely on constructing a synchronized product between the task automaton and the full system model, leading to prohibitive computational complexity (*55*). To address this double-exponential explosion, numerous methods have been proposed such as sampling (*56*), hierarchical abstraction (*57*), and partial-order analyses (*58*). More critically, they are generally unsuitable for open worlds where new tasks can be triggered online, as this would require frequent and expensive re-computation. Consider $\mathcal{B}_{\varphi_k}$ associated with the mission specification $\varphi_k$, and there are $\mathcal{M}$ groups of robots denoted by $\mathcal{C}_m \subset \mathcal{N}$ and $\mathcal{M} \triangleq \{1, \cdots, M\}$. Denote by $\Gamma_m$ the local plan of group $\mathcal{C}_m$ as the sequence of tasks to be accomplished, for $m \in \mathcal{M}$. Our objective is to find local plans such that the resulting trace satisfies all $\varphi_k \in \Phi_t$ that satisfy the LTL specifications while ensuring complete task allocation and minimizing the maximum execution time across all groups. And the plan is required to induce a feasible accepting path in the corresponding Büchi automaton, thereby guaranteeing satisfaction of the temporal-logic requirements. In addition, the plan must respect resource-allocation constraints, i.e., every instantiated task is uniquely assigned to a robot group that possesses the requisite capabilities. Finally, the plan must satisfy temporal scheduling constraints, including task-precedence relations and non-negative task start times. Detailed optimization formulation is provided in Supplementary Methods.

To avoiding constructing the full product model, we propose a simultaneous mission reasoning and group assignment framework. The core of our approach is a multiple automaton guided tree search, expressed as:

$$\{\Gamma_m\}_{m \in \mathcal{M}}^{\star} \triangleq \texttt{MultiAutoGuidedSearch}\,(\{\mathcal{B}_{\varphi_k}\}_{k \in \mathcal{K}}, \{\mathcal{C}_m\}_{m \in \mathcal{M}}, \Phi_t), \qquad (2)$$

where the synthesized local plan must satisfy the LTL constraints. The search structure is organized as a tree $\mathfrak{T} \triangleq (\mathcal{V}, \to)$, where $\mathcal{V} \triangleq \{v\}$ is the set of nodes, and $\to \subset \mathcal{V} \times \mathcal{V}$ defines the edges. Moreover, each node $v \triangleq \left(\{\Gamma_m, m \in \mathcal{M}\}, \left\{\hat{Q}_k, k \in \mathcal{K}\right\}\right)$ consists of two terms: the first as the local plans of all groups $\{\mathcal{C}_m\}$ as a partial assignment, and the second as the set of current reachable states in $\mathcal{B}_{\varphi_k}$. Initially, the root node is given by $v_0 \triangleq \left(\varnothing, \left\{\hat{Q}_k^0, k \in \mathcal{K}\right\}\right)$. Then this solution is adopted in a receding horizon manner, namely, a set of $H > 0$ candidate nodes is selected for parallel expansion, i.e., $\mathcal{V}_H \triangleq \left\{v_1^*, \cdots, v_H^*\right\}$, where $v_h^* \triangleq \text{argmin}_{v \in \mathcal{V}}\{\chi(v)\}$ for $h = 1, \cdots, H$; and $\chi : \mathcal{V} \to \mathbb{R}^+$ is a value function defined as:

$$\chi(v) \triangleq \max_{m \in \mathcal{M}} T_m + \eta_1 \sum_{m \in \mathcal{M}} C_m + \eta_2 \sum_{k \in \mathcal{K}} \min_{q_k \in \hat{Q}_k} \psi\left(q_k, Q_F^k\right) \tag{3}$$

Where $\eta_1, \eta_2 > 0$ are weighting parameters; $T_m > 0$ is the ending time for local plan $\Gamma_m$; $C_m > 0$ is the estimated cost of $\Gamma_m$; and $\psi\left(q_k, Q_F^k\right) > 0$ returns the length of the shortest path from state $q_k$ to any final state within $Q_F^k$. Namely, the first term measures the makespan of the current assignments; the second term estimates the overall cost of all assigned tasks; and the third term takes into account the overall progress of each mission. The complete mathematical definitions of all constraints and the detailed algorithm are provided in the Supplementary Methods section "Subtask Generation and Group Assignment via Multiple Automaton Guided Search". Once a feasible assignment is obtained, the temporal dependencies among tasks are extracted and visualized as a DAG, building upon our prior work (*11*). More details of DAG are found in the “Directed Acyclic Graphs of Tasks” section under the Supplementary Method.

**Subtask Generation**

Given the local plan $\Gamma_m$ and its assigned tasks $\Omega_m \triangleq \{\omega_1, \cdots, \omega_{L_m}\}$ for a robot group $\mathcal{C}_m$, each task $\omega \in \Omega_m$ is not directly executable but requires decomposition into a sequence of subtasks and the allocation of associated resources. In open and unknown environments, both the feasible subtask sequences and the distribution of required resources are uncertain. Traditional methods relying on manually designed guidelines are inflexible and domain-specific. To address this, we propose a task reasoning method based on LLMs. For each task $\omega$, a structured prompt $\Pi_\omega$ is designed to encapsulate the current context:

$$\Pi_\omega \triangleq \texttt{PromptDesign}(\omega, \mathcal{K}_t, \mathcal{A}_m, \Xi_t), \tag{4}$$

where $\mathcal{K}_t$ is the knowledge base (e.g., maps, symbolic rules), $\mathcal{A}_m$ describes the group's capabilities, and $\Xi_t$ contains structured multi-modal perceptions. Unlike one-shot LLM inference (*58*), we adopt the multi-stage reasoning to generate subtask sequences:

$$\mathcal{G}_\omega \triangleq \texttt{MultiStageLLM}(\Pi_\omega), \tag{5}$$

our method employs a RAG framework, first retrieving relevant subtask sequences from a plan library to provide contextual grounding, followed by a three-stage reasoning process: (I) domain knowledge analysis, (II) subtask guidance focusing on complex constraints, and (III) subtask sequencing. This enhances the consistency and feasibility of the generated plans. The LLM then outputs a set of labeled subtask graphs called layered DAGs, denoted by $\mathcal{G}_\omega \triangleq (\mathcal{S}_\omega, \mathcal{D}_\omega, \mathcal{L}_\omega)$ with $\mathcal{S}_\omega$ being the subtask nodes, $\mathcal{D}_\omega$ being the precedence relations and labels $\mathcal{L}_\omega$ specifying resource needs and execution attributes. If a resource required by a subtask $\sigma \in S_\omega$ is missing, an exploration subtask is inserted into the graph, i.e., $\mathcal{G}_\omega \triangleq \mathcal{G}_\omega \cup \{(\sigma_r^{\text{exp}}, \ell_r) | r \in \mathcal{R}_\omega^{\text{miss}}\}$, where $\mathcal{R}_\omega^{\text{miss}}$ is the set of missing resources, and each exploration task $\sigma_r^{\text{exp}}$ is labeled $\ell_r$ with potential locations. The resulting subtask graph denoted by $\mathcal{G}_\omega$ thus includes both the feasible subtask sequence and the necessary exploration actions to acquire unavailable resources. The detailed prompt design is found in the section of "LLM-assisted Subtask Generation with RAG" in Supplementary Methods.

## Uncertainty-aware Subtask Assignment

After obtaining multiple candidate subtask sequences for each task, the next step is to select a specific sequence for each task and assign its subtasks to the robots in group $\mathcal{C}_m$. This must satisfy temporal constraints such as precedence and priority, while balancing two objectives, i.e., minimizing the local makespan and bounding the uncertainty introduced by probabilistic exploration subtasks, which is solved as follows:

$$\Gamma_m^{\text{sub}}(t) \triangleq \texttt{OnlineAssign}(\mathcal{G}_\omega, \mathcal{A}_m^{\text{sub}}, \{(\hat{p}_\sigma, d_\sigma)\}, \varepsilon_m), \tag{6}$$

where $\hat{p}_\sigma$ and $d_\sigma$ represent the success probability and duration of subtask $\sigma$, $\varepsilon_m \in (0,1)$ is a permissible uncertainty bound and $\mathcal{A}_m^{sub}$ represents the capability of group $\mathcal{C}_m$. We address this through a parallel evaluation framework. Candidate subtask sequences are enumerated and assessed concurrently based on temporal feasibility, resource requirements, and dispatch efficiency to robots. This can be readily formulated as a bounded-uncertainty mixed-integer linear program (MILP) over the assignment variables and the schedule of the subtasks, which are solved by existing solvers. The combination yielding the minimum objective $\Gamma_m^{\text{sub}}$ is selected. Dispatch is performed in a rolling-horizon manner, i.e., after a certain number of subtasks is executed, the eligibility set is updated and the MILP is resolved. This ensures adaptability to dynamic progress and resource availability. Further details are provided in the section of "Subtask Allocation via MILP with Uncertainty Handling" under Supplementary Methods.

## Human-swarm Collaboration

The human–swarm collaboration layer combines sparse human verification with automated execution monitoring to improve transparency and reliability during long-horizon missions. Human operators remain in the loop to validate intermediate planning outputs and provide corrective guidance when needed, whereas an automated monitoring module tracks execution progress and checks compliance with temporal and synchronization constraints in real time. Together, these mechanisms support interpretable planning and online supervision without requiring continuous low-level control from the operator.

### *Human Interaction*

To improve transparency and reduce the impact of perception or planning errors, the system incorporates human-in-the-loop verification throughout the planning pipeline. Human validation is applied hierarchically, from semantic interpretation of environmental features to task decomposition and final subtask allocation, so that operators can correct potential perception errors, refine LLM-generated outputs, and adjust assignments when needed. Specifically, as shown in Fig. 2E, we develop four types of human-swarm interaction interfaces through the design of interactive UI elements: (i) It supports the label editing, allowing operators to modify feature recognition results via a trained YOLO-v7 model during context analysis. (ii) It enables interaction with LLMs, providing both text and voice input for real-time communication to guide and confirm the LLMs' outputs. (iii) It supports drag-and-drop functionality for task and subtask assignment, enabling operators to allocate tasks or subtasks to robots based on their own needs and expertise. (iv) We include a drawing-based interface that allows operators to manually define exploration areas for the robot, enhancing the efficiency of its exploration.

### *Execution Monitoring with Correctness Guarantee*

To monitor whether execution remains consistent with the temporal and synchronization requirements of the task specification, we combine two complementary mechanisms: (I) Temporal Logic Monitoring tracks the complete task execution trajectory via NBA that verify task completion against LTL formulas by an automaton visualization, and (II) Dynamic Synchronization via DAGs that enforce inter-task temporal constraints and synchronization for collaborative task execution. The monitoring mechanism updates the reachable states as:

$$\hat{Q}_k(t+1) = \{q' \in Q^k \mid \exists q \in \hat{Q}_k(t), \sigma \in \Sigma^k : (q, \sigma, q') \in \delta^k\}, \tag{7}$$

ensuring repeated visits to accepting states, i.e., ensuring that the distance to the accepting states $\sum_{k \in \mathcal{K}} \min_{q_k \in \hat{Q}_k} \psi\left(q_k, Q_F^k\right)$ continuously decreases to zero. If the distance becomes infinite, it indicates the absence of a path to the accepting states. Concurrently, precedence and synchronization constraints are dynamically enforced via:

$$\max_{i \in \mathcal{C}_{\omega_{m_2}}} s_m^i \geq \min_{j \in \mathcal{C}_{\omega_{m_1}}} e_m^j, \tag{8}$$

$$s_{m_1}^i = s_{m_2}^j, \forall i \in \mathcal{C}_{\omega_{m_1}}, j \in \mathcal{C}_{\omega_{m_2}}, \tag{9}$$

where $s_m^i$ is the start time and end time $e_m^j$ of performing task $\omega_m$. These mechanisms provide provable correctness guarantees while adapting to the evolution and new tasks during the execution. Further details are provided in the section of "Human-in-the-Loop Verification, Monitoring and Online Adaptation" under Supplementary Methods.

### Online Adaptation Mechanism

To support execution in dynamic environments, the system employs an online adaptation mechanism. Rather than reprocessing the entire pipeline, specific modules are triggered in response to changes such as new task types, environmental updates, task instance variations, resource discoveries, or robot failures. Key adaptations are proposed as follows: (i) New Task Type: When new tasks are introduced, the cloud-layer mission reasoning module generates new LTL specifications and automata. These are assigned to appropriate robot groups, with the edge layer decomposing tasks into subtasks for redistribution. (ii) New Task Instances: For new instances of existing tasks (e.g., newly detected survivors), the cloud layer allocates the task, and the edge layer redistributes subtasks to robots within the group. (iii) New Resource Type: Novel environmental features (e.g., water alongside a fire-extinguishing task) prompt the subtask generation module to update the layered DAG, which is forwarded to the subtask assignment module for revision and synchronization with the cloud database. (iv) New Resource Instances: When known resources (e.g., a new water reservoir) are detected, the system updates resource information and redistributes tasks accordingly. (v) Robot Failure: In case of robot failure, the subtask assignment module recalculates feasible plans for remaining robots, ensuring continued operation despite the failure. This modular, multi-level adaptation process ensures that the system maintains operational efficiency, even in unpredictable and dynamic conditions. The detailed description is displayed in the "Multi-level Online Adaptation" section of the Supplementary Methods.

## Data availability

All data generated in the simulation and hardware experiments, together with all questionnaires used in the human–computer interaction experiments, are available from Zenodo at https://doi.org/10.5281/zenodo.19014493.

## Code availability

The code developed in this study is available via GitHub at https://trio-pku.github.io/DEXTER-LLM-Plus/.

**Acknowledgments:** We express our gratitude to Y. Deng, H. B. Sun, A. M. Li, Q. Shi, W. Li and K. Liu for the invaluable suggestions on the manuscript; Y. H. Luo and S. Zhang for the contribution in robot hardware debugging and for providing photography and audio recording services; and J. S. Wei, C. Wang, H. Z. Dong, and Y. Y. Zhang for the assistance in physical experiments. Our heartfelt thanks go out to L. Bai, B. C. Yao, and P. X. Shu from the Xiwang Dong Research Group at Beihang University for the support in terms of venue. We also thank the 18 volunteers for providing feedback while using our system, which served as the source for our data analysis.

**Author contributions:** J. F. Chen and Y. X. Zhu participated in the design of the entire algorithm, software, and hardware, conducted all experimental tests, and contributed to manuscript writing. A. Zhuo and S. Zhang were involved in the debugging of the swarm, including the navigations and localization algorithms for UAVs, legged robots, and wheeled vehicles. X. T. Zhang conducted the integration of the physical experiments and data collection. G. H. Wen and X. W. Dong provided valuable insights into the experimental procedures and manuscript. M. Guo and Z. K. Li supervised the research, provided financial support and hardware equipment, contributed to algorithmic discussions, and participated in manuscript revision.

**Funding:** Include This work was supported by the National Natural Science Foundation of China under grant no. U2241214, T2121002, and 62373008.

**Competing interests:** The authors declare that they have no competing interests.

# Supplementary Materials for
# Melding LLM and temporal logic for reliable human-swarm collaboration in complex environments

Junfeng Chen, *et al.*
[*]Corresponding author:
Meng Guo Email: meng.guo@pku.edu.cn
Zhongkui Li Email: zhongkli@pku.edu.cn

**This PDF file includes:**

# Supplementary Methods

## *Simultaneous Task Decomposition and Group Assignment*

This section offers a comprehensive overview of Linear Temporal Logic (LTL) fundamentals and strategies for task decomposition. Specifically, we introduce a novel method that integrates LTL with nondeterministic Büchi automata (NBA) and a multiple automation guided tree search method to achieve efficient task allocation. A concrete example is provided to illustrate the proposed approach.

### Preliminary of Linear Temporal Logic

The tasks $\omega_m$ are nested following the syntax of LTL (*33*), e.g., via $\varphi \triangleq \top | p | \varphi_1 \wedge \varphi_2 | \neg\varphi | \bigcirc\varphi | \varphi_1 U \varphi_2$ where $\top \triangleq \text{True}$, $p \in AP$, $\bigcirc$ (next), $U$ (until) and $\bot \triangleq \neg\top$. or other derived operators like $\square$(always), $\Diamond$(eventually), $\Rightarrow$(implication). The full semantics and syntax of syntactic co-safe LTL (sc-LTL) are omitted here for brevity, see e.g., (*40*). An infinite word $w$ over the alphabet $2^{AP}$ is defined as an infinite sequence $W = \sigma_1\sigma_2\cdots, \sigma_i \in 2^{AP}$. The language of $\varphi$ is defined as the set of words that satisfy $\varphi$, namely, $\mathcal{L} = \text{Words}(\varphi) = \{W \mid W \vDash \varphi\}$ and $\vDash$ is the satisfaction relation. However, there is a special class of LTL formula called *co-safe* formulas, which can be satisfied by a set of finite sequence of words. They only contain the temporal operators $\bigcirc$, $U$ and $\Diamond$ and are written in positive normal form.

Given an LTL formula $\varphi$ mentioned above, the associated NBA can be derived with the following structure. A NBA $\mathcal{B}$ is a 5-tuple: $\mathcal{B} = (Q, Q_0, \Sigma, \delta, Q_F)$, where $Q$ is the set of states; $Q_0 \subseteq Q$ is the set of initial states; $\Sigma = AP$ is the allowed alphabet; $\delta : Q \times \Sigma \to 2^Q$ is the transition relation; $Q_F \subseteq Q$ is the set of *accepting* states.

Given an infinite word $w = \sigma_1\sigma_2\cdots$, the resulting *run* within $\mathcal{B}$ is an infinite sequence $\rho = q_0q_1q_2\cdots$ such that $q_0 \in Q_0$, and $q_{i+1} \in \delta(q_i, \sigma_i)$ hold for all index $i \geq 0$. A run is called *accepting* if it holds that $\inf(\rho) \cap Q_F \neq \varnothing$, where $\inf(\rho)$ is the set of states that appear in $\rho$ infinitely often. In general, an accepting run can be written in the prefix-suffix structure, where $q_0$ the prefix starts from an initial state and ends in an accepting state; and the suffix is a cyclic path that contains the same accepting state. It is easy to show that such a run is accepting if the prefix is repeated once, while the suffix is repeated infinitely often. In general, the size of $\mathcal{B}$ is double exponential to the length of formula $\varphi$.

### Task Specification and Comprehension under LTL Specifications

As described in the main text, we utilize the plan $\mathcal{K}$ to specify the task specification. Below, we present a comprehensive example in JSON format, detailing specific event types, corresponding tasks, temporal constraints among tasks, and executable subtask sequences for each task:

```
[
  "When encountering an alkane gas flame, first carefully inspect the
source and surrounding environment to assess risks. To extinguish the
fire, consider two approaches: shutting off relevant valves to cut off
```

```
the gas supply or using water spray to cool and suppress the flame.
Finally, monitor the fire to track its development and ensure complete
extinguishment.",
  "For high-temperature liquid flames, begin by inspecting the scene to
assess the extent and potential spread. One method involves using
asbestos felt to cover the flames, thereby cutting off the oxygen
supply. If unavailable, spray water to cool the area. Finally, monitor
to ensure complete extinguishment and no residual hazards.",
  "For high-voltage electrical flames, careful inspection is paramount.
Then, a recommended approach is to operate switches to cut off the power
supply, mitigating electric shock and fire spread risks. Alternatively,
use foam spray to extinguish the flame. Throughout, continuously monitor
to ensure safety and complete extinguishment.",
  "When rescuing trapped individuals, first inspect the scene to locate
and assess the victim's condition. Then, clear debris or hazards to
establish safe rescue conditions. Next, carefully rescue the person.
Finally, monitor the victim's condition and provide support until safe
rescue is completed.",
  "When handling poisoned individuals, first inspect the scene to
identify the toxin type and the person's condition. Then, administer
oxygen to prevent asphyxiation. Next, carefully rescue them. Finally,
monitor their recovery.",
  "In the event of hydrogen sulfide leakage, immediately inspect the
spread range. If feasible, safely ignite the gas to control the hazard,
then apply solid sprays such as activated carbon for adsorption.
Continuously monitor the site to minimize residual hazards.",
  "For storage tank protection, first carefully check the tank's status
to obtain key information. Then, apply water to cool and shield the
tank, followed by necessary repairs. Finally, monitor for damage or
leaks."
]
```

Additionally, beyond this JSON representation, our framework supports alternative plan formats, e.g., text, yaml, or XML, provided that they encompass the essential components outlined above. Building on these plans, our experts formalize the LTL formulas for chemical plant emergency response tasks based on their understanding of the aforementioned plans and the foundational LTL formulas presented as follows:

$$
\begin{aligned}
\phi = {} & (\square\Diamond \text{monitor}) \\
& \wedge \square\Big( ((\text{tp} \vee \text{poi}) \rightarrow \Diamond(\text{insp} \wedge \Diamond \text{res})) \\
& \wedge ((\text{tp} \vee \text{poi})\,\mathcal{U}(\text{res} \wedge \text{monitor})) \\
& \wedge (\text{af} \rightarrow \Diamond(\text{insp} \wedge \Diamond(\text{cut} \vee \text{cool}) \wedge \Diamond \text{ext})) \\
& \wedge (\text{htlf} \rightarrow (\text{af}\,\mathcal{U}(\text{insp} \wedge \Diamond(\text{cover} \vee \text{cool}) \wedge \Diamond \text{ext}))) \\
& \wedge (\text{hvf} \rightarrow (\text{insp} \wedge \Diamond(\text{cut} \wedge \Diamond \text{ext}))) \\
& \wedge (\text{h2s} \rightarrow (\text{insp} \wedge \Diamond(\text{ignite} \vee \text{ads}) \wedge \Diamond \text{monitor})) \\
& \wedge (\text{tank} \rightarrow (\text{insp} \wedge \Diamond(\text{cool} \wedge \Diamond(\text{repair} \wedge \text{monitor})))) \Big)
\end{aligned}
\tag{S1}
$$

The LTL formulation in (S1) not only specifies temporal constraints between tasks but also encodes the executable subtask sequences defined in the emergency plans. The abbreviated terms (detailed in Table S1) systematically map to both task categories and concrete response actions. Crucially, this formalization implements a hierarchical priority structure: (I) Rescue operations $(\text{tp}, \text{poi})$ must precede all other tasks; (II) Alkane flame mitigation $(\text{af})$ takes precedence over high-temperature liquid fire handling (htlf); (III) Remaining tasks (hvf, h2s, tank) may execute concurrently. This priority schema ensures critical life-saving interventions receive immediate attention while maintaining safe parallelism for less interdependent hazards

**Table S1**: Abbreviations used in the LTL formula and their corresponding meanings.

| Abbreviation | Meaning | Abbreviation | Meaning |
|---|---|---|---|
| **tp** | Trapped person | res | Rescue |
| **poi** | Poisoned person | cut | Cut power/shut valve |
| **af** | Alkane flame | cool | Water spray/cool tank |
| **htlf** | High-temp liquid flame | cover | Cover asbestos |
| **hvf** | High-voltage flame | ext | Extinguish |
| **h2s** | $H_2S$ leakage | ignite | Ignite gas |
| **tank** | Tank damage | ads | Adsorb carbon |
| **insp** | Inspect | repair | Repair tank |
| **monitor** | Monitor | | |

Building on the LTL formulation in (S1), we employ $LTL2BA$ toolbox described in the main text to encode into the corresponding NBA, as depicted in Fig. S1. Following our established methodology, this NBA is algorithmically parsed to extract a partial order set (poset) of mission tasks, as seen in Fig. S2. The poset explicitly captures the essential precedence constraints and concurrency relationships implied by the temporal logic and priority scheme. This partial order is subsequently visualized as a directed acyclic graph (DAG) as shown in Fig. S3, where nodes represent atomic tasks and directed edges denote precedence requirements. The DAG structure directly implements the hierarchical priorities defined in the LTL formula: rescue operations (tp, poi) dominate all other tasks; alkane flame mitigation (af) precedes high-temperature liquid fire response (htlf); while high-voltage fires (hvf), hydrogen sulfide leaks (h2s), and tank protection (tank) operates in concurrent branches. This representation provides both a formal verification artifact and an executable task framework for emergency response coordination.

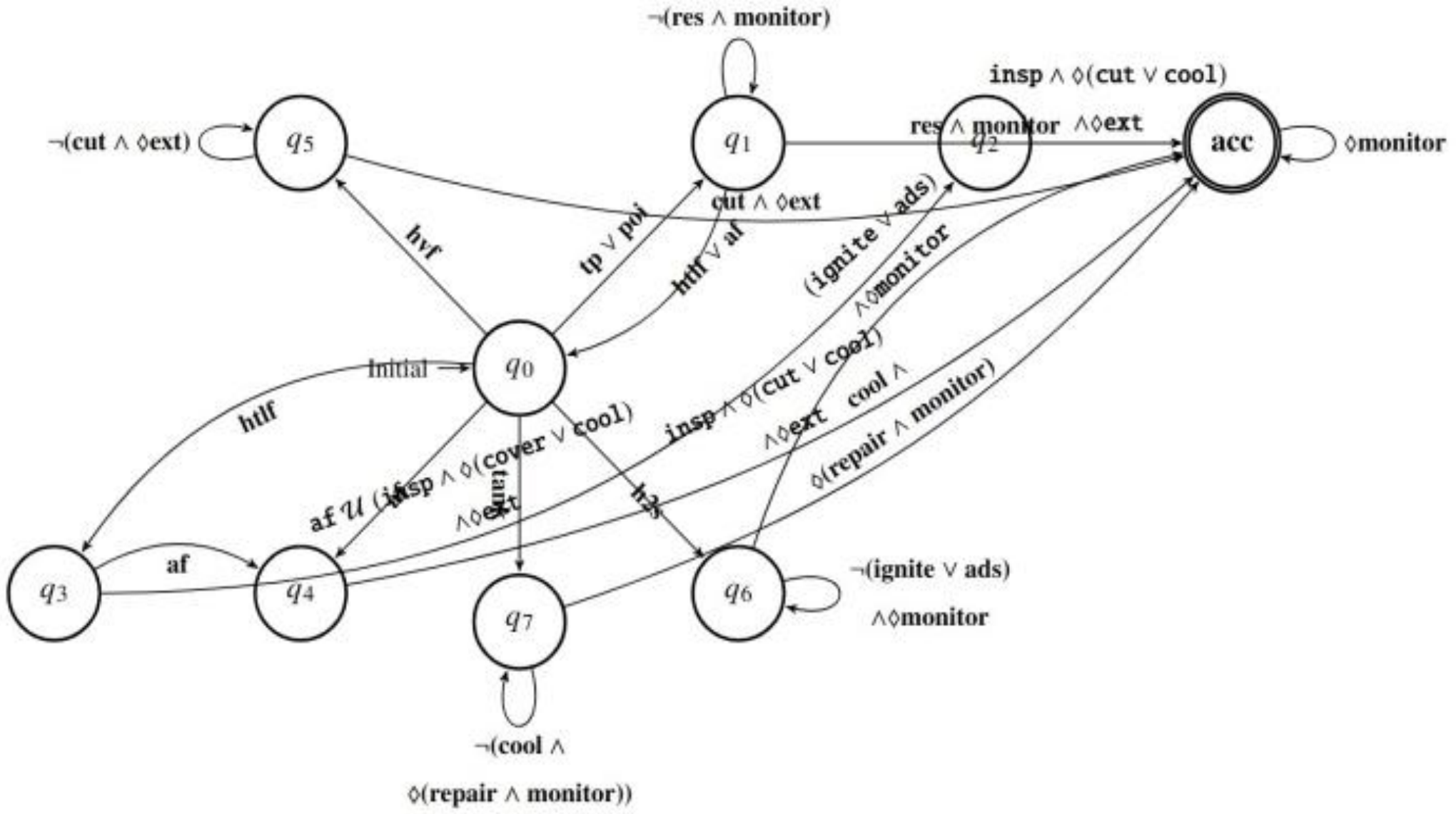

**Fig. S1:** The automaton corresponding to Eq. S1 above.

**R-poset** $\mathcal{P}_\varphi = (\Omega_\varphi, \preceq_\varphi, \neq_\varphi)$

$\Omega_\varphi = \{\mathtt{tp}, \mathtt{poi}, \mathtt{af}, \mathtt{htlf}, \mathtt{hvf}, \mathtt{h2s}, \mathtt{tank}\}$

$\preceq_\varphi = \{(\mathtt{tp}, \omega), (\mathtt{poi}, \omega)\ \forall \omega \in \mathcal{H},\ (\mathtt{af}, \mathtt{htlf})\}$

$\neq_\varphi = \{(\alpha, \beta) \mid \alpha \in \mathcal{R}, \beta \in \mathcal{H}\}$

where $\mathcal{R} = \{\mathtt{tp}, \mathtt{poi}\}$,

$\mathcal{H} = \{\mathtt{af}, \mathtt{htlf}, \mathtt{hvf}, \mathtt{h2s}, \mathtt{tank}\}$

**Fig. S2**: Formal R-poset specification derived from LTL formula (S1).

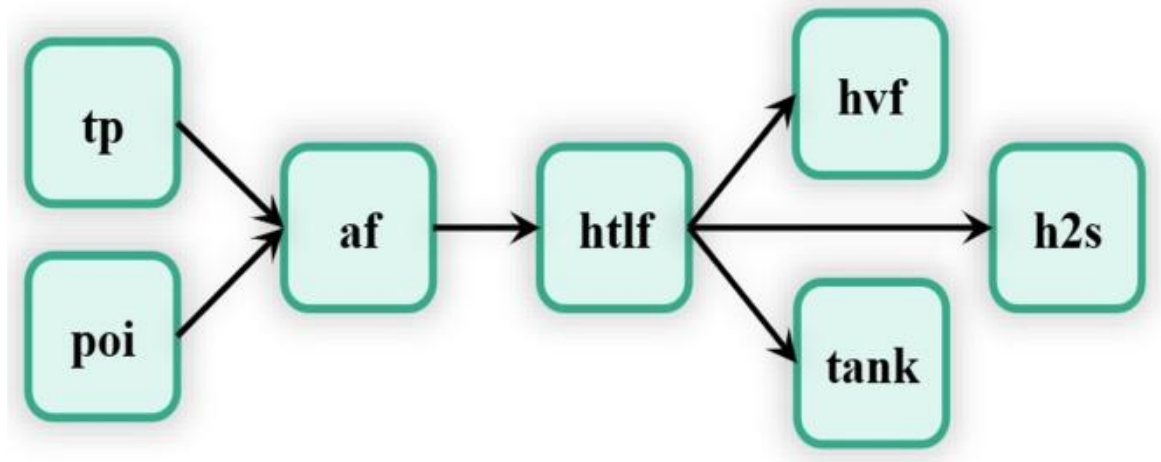

**Fig. S3**: DAG representation of mission tasks derived from the R-poset. Arrows indicate precedence constraints between atomic tasks.

## *Subtask Generation and Group Assignment via Multiple Automaton Guided Search*

As mentioned in the main text, the simultaneous optimization of task decomposition and group assignment is formulated as a multi-objective optimization problem as follows:

$$\min_{\{\Gamma_m\}_{m\in\mathcal{M}}} \max_{m\in\mathcal{M}} T_m \tag{S2}$$

$$s.t. \exists \pi_k \text{ with } q_0 \in Q_0^k, q_n \in Q_F^k, \forall k \in K \tag{S3}$$

$$\sum_{m\in\mathcal{M}} I\left[\omega \in \Gamma_m\right] = 1, \forall \omega \in \bigcup_{k\in\mathcal{K}} \Sigma^k \tag{S4}$$

$$\omega \in \mathcal{A}_m, \forall m \in M, \forall \omega \in \Gamma_m \tag{S5}$$

$$S_{\omega'} \geq S_\omega + d_\omega, \forall m \in M, \forall \omega, \omega' \in \Gamma_m \text{ with } \omega \prec \omega' \tag{S6}$$

$$T_m \geq S_\omega + d_\omega, \forall m \in M, \forall \omega \in \Gamma_m \tag{S7}$$

$$S_\omega \geq 0, T_m \geq 0, \forall \omega \in \Omega, \forall m \in M \tag{S8}$$

Then we propose a multiple automaton guided tree search method to solve the above problem as follows:

$$\{\Gamma_m\}^\star_{m\in\mathcal{M}} \triangleq \text{MultiAutoGuidedSearch}\left(\{\mathcal{B}_{\varphi_k}\}_{k\in\mathcal{K}}, \{\mathcal{C}_m\}_{m\in\mathcal{M}}, \Phi_t\right) \tag{S9}$$

More specifically, consider the NBA $\mathcal{B}_{\varphi_k} \triangleq \left(Q^k, Q_0^k, \Sigma^k, \delta^k, Q_F^k\right)$ associated with the mission specification $\varphi_k \in \Phi_t$, where $\Phi_t \triangleq \{\varphi_1, \cdots, \varphi_K\}$ is the set of missions known at time $t > 0$ and $\mathcal{K} \triangleq \{1, \cdots, K\}$. Moreover, there are $M$ groups of robots denoted by $\mathcal{C}_m \subset \mathcal{N}$ and $\mathcal{M} \triangleq \{1, \cdots, M\}$. Denote by $\Gamma_m$ the local plan of group $\mathcal{C}_m$. as the sequence of tasks to be accomplished, for $m \in \mathcal{M}$.

The search structure is organized as a tree $\mathfrak{T} \triangleq (\mathcal{V}, \to)$, where $\mathcal{V} \triangleq \{v\}$ is the set of nodes, and $\to \subset \mathcal{V} \times \mathcal{V}$ defines the edges. Moreover, each node $v \triangleq \left(\{\Gamma_m, m \in \mathcal{M}\}, \{\hat{Q}_k, k \in \mathcal{K}\}\right)$ consists of two terms: the first as the local plans of all groups $\{\mathcal{C}_m\}$ as a partial assignment, and the second as the set of current reachable states in $\mathcal{B}_{\varphi_k}$ for each mission $\varphi_k \in \Phi_t$. Initially, the root node is given by $v_0 \triangleq \left(\varnothing, \{\hat{Q}_k^0, k \in \mathcal{K}\}\right)$. Then, the search proceeds through four primary stages:

*(I) Selection.* A set of $H > 0$ candidate nodes is selected for parallel expansion, i.e., $\mathcal{V}_H \triangleq \{v_1^\star, \cdots, v_H^\star\}$, where $v_h^\star \triangleq \mathbf{argmin}_{v\in\mathcal{V}} \{\chi(v)\}$ for $h = 1, \cdots, H$; and $\chi : \mathcal{V} \to \mathbb{R}^+$ is a value function defined as:

$$\chi(v) \triangleq \max_{m\in\mathcal{M}} T_m + \eta_1 \sum_{m\in\mathcal{M}} C_m + \eta_2 \sum_{k\in\mathcal{K}} \min_{q_k\in\hat{Q}_k} \psi\left(q_k \;\; Q_F^k\right) \tag{S10}$$

where $\eta_1, \eta_2 > 0$ are weighting parameters; $T_m > 0$ is the ending time for local plan $\Gamma_m$; $C_m > 0$ is the estimated cost of $\Gamma_m$; and $\psi\left(q_k, Q_F^k\right) > 0$ returns the length of the shortest path from state $q_k$ to *any* final state within $Q_F^k$. Namely, the first term in (S10) measures the makespan of the current assignments; the second term estimates the overall cost of all assigned tasks; and the third term takes into account the overall progress of each mission.

*(II) Expansion.* Each selected node $v_h^\star \in \mathcal{V}_H$ is expanded by assigning an additional task to one group of robots. Consider the set of candidate tasks for group $\mathcal{C}_m$, denoted by

$$\Omega^-_{v_h^\star, m} \triangleq \{\omega \in \left(\Sigma^k \cap \mathcal{A}_m\right) \mid \exists q_k \in \hat{Q}_k : \delta^k\left(q_k, \omega\right) \in Q^k\} \tag{S11}$$

where $\omega$ is an atomic task symbol from the alphabet $\Sigma^k$, $\delta^k : Q^k \times \Sigma^k \to 2^{Q^k}$ is the transition function of the NBA $\mathcal{B}_{\varphi_k}$, and $\mathcal{A}_m \triangleq \{\omega \in \Sigma^k \mid \mathcal{C}_m \text{ can execute } \omega\}$ is the set of tasks feasible for group $\mathcal{C}_m$ given its capabilities. Hence, a candidate task must both enable a valid state transition and be executable by the selected group.

For each candidate task $\omega \in \Omega^-_{v_h^\star, m}$, a child node can be created by augmenting the local plan $\Gamma_m$ of group $\mathcal{C}_m$ with $\omega$, i.e.,

$$v^+ \triangleq (\{\Gamma_1, \cdots, \Gamma_m^+, \cdots, \Gamma_M\}, \{\hat{Q}_k^+\}_{k \in \mathcal{K}}) \tag{S12}$$

where $\Gamma_m^+$ is obtained by appending $\omega$ to $\Gamma_m$; the set of tasks $\Omega_m$ within $\Gamma_m$ is also updated by adding $\omega$; and $\hat{Q}_k^+ \triangleq \{\delta^k(q_k, \omega) \mid q_k \in \hat{Q}_k\}$ is the updated set of reachable states in the automaton for mission $\varphi_k$. Consequently, the edge $(v_h^\star, v^+)$ is inserted into the search tree $\mathfrak{T}$. If no valid and feasible transition can be realized, then $\Omega^-_{v_h^\star, m} = \varnothing$ and the node cannot be expanded. Lastly, given the partial local plans $\{\Gamma_m\}_{m \in \mathcal{M}}$, a scheduling problem is formulated to coordinate the execution times of tasks among different groups. Specifically, a linear program is defined over the starting and ending times of each task, subject to navigation and coordination constraints imposed by the local modules of each group, with the objective of minimizing the overall makespan.

*(III) Bounding.* To reduce unnecessary exploration, each node is evaluated through a *performance profile* that retains detailed information about the plans of all groups and the progress of all missions, i.e.,

$$\zeta(v) \triangleq [\{T_m\}_{m \in \mathcal{M}}, \{C_m\}_{m \in \mathcal{M}}, \{\min_{q_k \in \hat{Q}_k} \psi(q_k, Q_F^k)\}_{k \in \mathcal{K}}] \tag{S13}$$

where $T_m$ is the ending time of the local plan $\Gamma_m$; $C_m$ is the estimated cost of $\Omega_m$; and the last term measures the minimum distance from the current automaton states $\hat{Q}_k$ to the accepting set $Q_F^k$ for each mission $\varphi_k$. Note that profile $\zeta(v)$ has dimension $(2M + K)$ and is non-negative.

Given two nodes $v_1$ and $v_2$, node $v_1$ is said to *dominate* node $v_2$ if

$$\zeta(v_1) \leq \zeta(v_2), \text{and } v_1 \neq v_2 \tag{S14}$$

where the inequality is understood element-wise across the entire vector in (S13). That is, node $v_1$ has no larger makespan or cost for any group, and no less mission progress for any specification, with strict improvement in at least one entry. In this case, node $v_2$ is marked as dominated and excluded from the set of nodes to be expanded. The set of all non-dominated nodes is called the frontiers, defined as

$$\overline{\mathcal{V}} \triangleq \{v \in V \mid \nexists v' \in V : v' \text{dominates } v\} \tag{S15}$$

which is updated each time a new node is added to the tree. This bounding procedure thus maintains only $\bar{\mathcal{V}}$ as candidates for further expansion, ensuring that strictly inferior nodes are pruned away and the search remains focused on promising branches of the assignment tree $\mathfrak{T}$.

*(IV) Termination.* The stages of selection, expansion, and bounding are repeated until the computation budget is exhausted or no new non-dominated nodes emerge. The current frontier is given by $\bar{\mathcal{V}}$ in (S15), and the subset of complete assignments is defined as

$$\bar{\mathcal{V}}^{\star} \triangleq \{v \in \bar{\mathcal{V}} \mid \left(\hat{Q}_k \cap Q_F^k\right) \neq \varnothing, \forall k \in K\} \tag{S16}$$

where the set of accepting states has been reached for all missions. Then, the incumbent node is then obtained as:

$$\bar{v}^{\star} \triangleq \mathbf{argmin}_{v \in \bar{\mathcal{V}}^{\star}} \left\{\chi\left(v\right)\right\} \tag{S17}$$

with $\chi\left(v\right)$ from (S10), and the global assignment is specified by the resulting plans $\{\Gamma_m\}_{m \in \mathcal{M}}$ together with execution times determined by the linear program introduced in Stage (II).

This framework offers several advantages over existing approaches. First, it circumvents the explicit construction of the synchronous product between the Büchi automaton, the local robot models as finite transition systems, and their global product, which is prohibitively large in multi-robot settings. Second, the search is anytime in the sense that it can return feasible intermediate solutions, and it is complete under the standard assumptions of finite branching and sufficient search budget. Third, it is well suited to partially known and dynamic environments where missions are triggered online, since a new specification can be incorporated simply by introducing its reachable state set $\hat{Q}_k$ without interfering with existing missions. Finally, the stages of node selection and expansion can be carried out in parallel, enabling efficient scaling to large teams and complex mission sets.

## *Directed Acyclic Graphs of Tasks*

As described in the main text, mission-level reasoning and group assignment are jointly formulated as a multi-objective optimization problem via multiple automaton guided tree search, we employ a relaxed partially ordered set (R-poset) introduced in our prior work (*31*) to capture the essential temporal constraints among tasks. Formally, the R-poset for specification $\phi$ is defined as a 3-tuple $\mathcal{P}_\phi = \left(\Omega_\phi, \prec_\phi, \neq_\phi\right)$, where (I) $\Omega_\phi$ denotes the set of tasks; (II) $\prec_\phi \subseteq \Omega_\phi \times \Omega_\phi$ represents the *precedence relation*: if $\left(\omega_h, \omega_\ell\right) \in \prec_\phi$, then $\omega_\ell$ can only commence after the initiation of $\omega_h$; (III) $\neq_\phi \subseteq \Omega_\phi \times \Omega_\phi$ defines the *exclusion relation*: if $\left(\omega_h, \omega_\ell\right) \in \neq_\phi$, then $\omega_h$ and $\omega_\ell$ are mutually exclusive and cannot execute concurrently. The complete family of R-posets $\mathcal{P}_\phi$ preserves temporal dependencies needed for scheduling, synchronization, and online monitoring, while being more compact than the original automaton structure. (*31*). Algorithmic particulars are omitted here for conciseness. For practical implementation, the R-poset is encoded as a DAG wherein nodes correspond to tasks, while labeled directed edges encode precedence and exclusion relations. Illustrative examples of mission comprehension via formal methods are provided in Fig. S3.

## LLM-assisted Subtask Generation with RAG

Our predefined plan library contains extensive prior knowledge, i.e., solutions to various tasks akin to general firefighting and rescue guidelines commonly employed across domains. Given the high-quality expertise embedded in this library, a key objective is to leverage its content by retrieving the most relevant prior cases for a given problem. These retrieved cases are subsequently integrated into the prompt of LLM to enhance its reasoning capabilities. To achieve this, each solution in the plan library is first converted into a string element within a list, structured as presented in the section of "Task Specification and Comprehension under LTL Specifications".

We then employ a sentence transformer model to quantify the semantic similarity between sentences, assigning a relevance score to each case. Specifically, the plan library is encoded into vector representations using the all-MiniLM-L12-V2 model (*59*). For a given task described in the natural language, the system retrieves the most semantically similar plan from the plan library, that is the case with the highest similarity score and then incorporates it as related knowledge into the LLM prompt. For instance, given the task "electrical fire," the RAG module retrieves the corresponding plan: "For high-voltage electrical flames, careful inspection is the highest priority. Then, one good way to extinguish it is operating switches to cut off the power supply, reducing the risk of electric shock and fire spread. Another way to extinguish the flame is using foam for liquid spray to extinguish the fire and use a metal net to prevent leakage of electricity. Throughout the process, repeatedly monitor to ensure safety and complete extinguishment." This retrieved knowledge is then appended to the prompt as the related knowledge for the LLM.

### *Design of Prompt*

Our approach employs a multi-stage reasoning procedure with LLMs, where the design of prompts for each stage is particularly critical. The prompt for every stage is constructed using a consistent core framework, which comprises instructions, robot description, and stage-specific elements. In the stage of "Related knowledge analysis", the prompt primarily takes input from the knowledge retrieved via RAG search based on the current contextual awareness, and outputs an analysis of the relevant information. This enables the LLM to enhance its capability in deriving feasible solutions based on the most pertinent subtask sequence. The stage of "Subtask Guide" integrates the analysis of feasible solutions and produces an intermediate representation of task dependencies in natural language. It yields a hierarchical task description that incorporates precedence constraints. Finally, in the stage of "Subtask Sequencing", the prompt incorporates both the subtask guide and the related knowledge analysis, generating a final subtask sequence represented as a layered DAG for human verification, along with a JSON-formatted subtask list to facilitate parsing by subsequent modules such as subtask allocation. Compared to one-shot inference methods such as deep thinking, this multi-stage prompt design enables the progressive refinement of the reasoning process. It enhances the inferential capability by emphasizing complex dependencies within the reasoning chain and grounding the output format, thereby improving the interpretability and accuracy. Some selected examples are illustrated in Fig. S4, Fig. S5 and Fig. S6.

**Task**: name: ...;
**Resources:** antidote, oxygen, valve, foam, water, asbestos_felt, activated_carbon
**Related Knowledge**: For alkane gas flames: inspect flame source/environment, then either: (1) shut off valves to cut gas supply, or (2) use water spray to cool/suppress flame. Finally monitor until fully extinguished.
**Instruction**: Combine skills to accomplish task. Output must be strict JSON format only.
**Robot Skills**:

- inspect: On-site investigation
- operate: Valve/switch control
- liquid_spray: Spray pressurized liquid
- monitor: Post-task observation

**Output Format**:

- related_knowledge_analysis: Analyze schemes/steps
- schemes_draft: Describe schemes in natural then formal language
- schemes:
    - Scheme_1:
    - steps:
        - Step 1:
            - required_skill: skill_name
            - required_resources: 0-1 nearby objects
            - dependency: prerequisite steps

**Fig. S4: Prompt outline for "Related Knowledge Analysis" stage.**

**Task**: …

**Related Knowledge**: ⋯

**Skills**: ⋯

**Instruction**: ⋯

**Related Knowledge Analysis**: ⋯

**Output Format**:

- schemes_draft: Use natural language to describe the schemes and steps involved. Then write the schemes formally in the following 'schemes' part. Note that there are only one scheme and one step is shown, but you can generate multiple schemes and steps.

**Fig. S5: Prompt outline for "Subtask Guide" stage.**

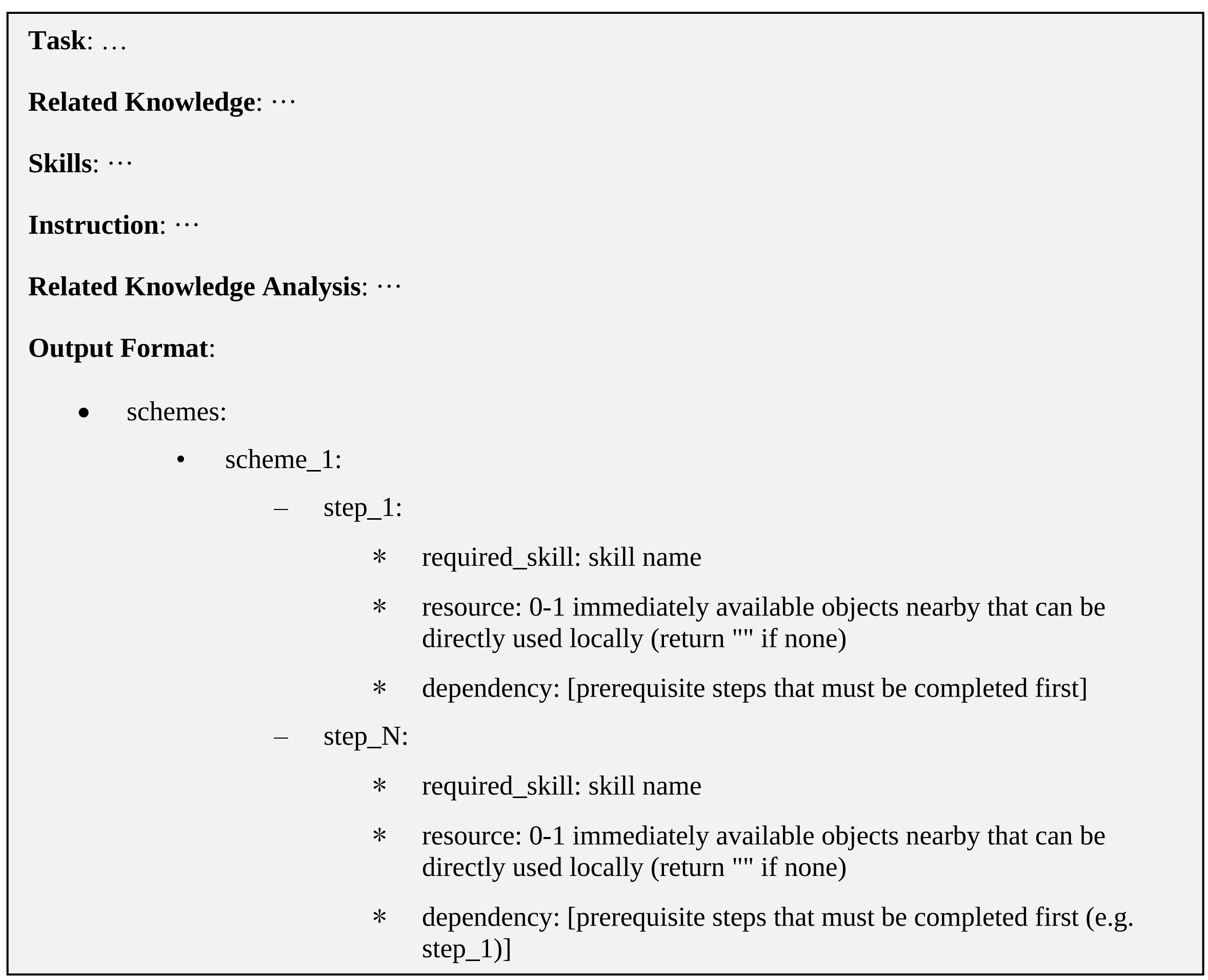

**Task**: …

**Related Knowledge**: ⋯

**Skills**: ⋯

**Instruction**: ⋯

**Related Knowledge Analysis**: ⋯

**Output Format**:

- schemes:
  - scheme_1:
    - step_1:
      - required_skill: skill name
      - resource: 0-1 immediately available objects nearby that can be directly used locally (return "" if none)
      - dependency: [prerequisite steps that must be completed first]
    - step_N:
      - required_skill: skill name
      - resource: 0-1 immediately available objects nearby that can be directly used locally (return "" if none)
      - dependency: [prerequisite steps that must be completed first (e.g. step_1)]

**Fig. S6: Prompt outline for "Subtask Sequencing" stage.**

## Subtask Allocation via MILP with Uncertainty Handling

As described in the main text, we employ a rolling horizon approach to dispatch a specific set of subtasks. For a given group of robots $\mathcal{C}_m$ and the set of candidate subtasks $\mathcal{E}_m(t) \triangleq \{\sigma_1, \cdots, \sigma_{|\mathcal{E}|}\}$ at time $t$, along with the group's capability constraints $\mathcal{A}_m^{sub}$, the success probability estimates $\hat{p}_\sigma$, duration estimates $d_\sigma$, and uncertainty budget $\varepsilon_m$ for each subtask, our optimization objective is to simultaneously minimize the local makespan while bounding the cumulative uncertainty within the specified budget. The dispatching optimization within each group is formulated as a mixed-integer linear programming problem as follows:

$$\min_{\Gamma_m^{\mathrm{sub}}(t)} \quad \max_{\sigma\in\Gamma_m^{\mathrm{sub}}(t)} (s_\sigma + d_\sigma) \tag{S18}$$

$$\textbf{s.t.} \quad \sum_{r\in\mathcal{C}_m} x_{r,\sigma} \le 1, \quad \forall \sigma \in \mathcal{E}_m(t) \tag{S19}$$

$$x_{r,\sigma} = 0, \quad \forall r \in \mathcal{C}_m, \forall \sigma \notin \mathcal{A}_m^{\mathrm{sub}}(r) \tag{S20}$$

$$S_{\sigma'} \ge S_\sigma + d_\sigma, \quad \forall \sigma, \sigma' \in \mathcal{E}_m(t) \text{ with } \sigma \prec \sigma' \tag{S21}$$

$$S_{\sigma'} \geq S_{\sigma} + d_{\sigma} - M(1 - y_{r,\sigma,\sigma'}), \quad \forall r \in \mathcal{C}_m, \forall \sigma, \sigma' \in \mathcal{E}_m(t) \tag{S22}$$

$$\sum_{\sigma \in \mathcal{E}_m(t)} (1 - \hat{p}_{\sigma}) \cdot \left( \sum_{r \in \mathcal{C}_m} x_{r,\sigma} \right) \leq \varepsilon_m \tag{S23}$$

$$\sum_{\sigma' \in \mathcal{E}_m(t)} y_{r,\sigma,\sigma'} = x_{r,\sigma}, \quad \forall r \in \mathcal{C}_m, \forall \sigma \in \mathcal{E}_m(t) \tag{S24}$$

$$y_{r,\sigma,\sigma'} + y_{r,\sigma',\sigma} \leq 1, \quad \forall r \in \mathcal{C}_m, \forall \sigma, \sigma' \in \mathcal{E}_m(t) \tag{S25}$$

$$s_{\sigma} \geq t, \quad \forall \sigma \in \mathcal{E}_m(t) \tag{S26}$$

$$x_{r,\sigma} \in \{0,1\}, \quad y_{r,\sigma,\sigma'} \in \{0,1\}, \quad s_{\sigma} \geq 0 \tag{S27}$$

where a binary variable $x_{r,\sigma}$ indicates whether robot $r \in \mathcal{C}_m$ is assigned to subtask $\sigma \in \mathcal{E}_m(t)$; $s_{\sigma} > 0$ represents the start time of subtask $\sigma$; and $y_{r,\sigma,\sigma'}$ denotes whether robot $r$ executes subtask $\sigma$ immediately before subtask $\sigma'$. The optimization objective (Eq. S18) minimizes the local makespan, defined as the maximum completion time among all assigned subtasks within the current horizon, ensuring efficient utilization of robotic resources and timely task completion. The subtask assignment constraint (Eq. S19) ensures that each subtask is assigned to at most one robot, preventing conflicting assignments. The capability constraint (Eq. S20) restricts task assignments based on robot capabilities, ensuring robots only execute subtasks that can be handled according to the capability matrix $\mathcal{A}_m^{sub}$. Temporal dependencies between subtasks are enforced through the precedence constraint (Eq. S21), which guarantees that if subtask $\sigma$ must precede subtask $\sigma'$ ($\sigma \prec \sigma'$), then $\sigma'$ cannot start before $\sigma$ completes. The robot scheduling constraint (Eq. S22) manages task sequencing for individual robots using big-M formulation, ensuring that when $y_{r,\sigma,\sigma'} = 1$, robot $r$ starts to perform the subtask $\sigma'$ only after completing subtask $\sigma$. Robustness against execution uncertainties is provided by the uncertainty budget constraint (Eq. S23), which bounds the cumulative uncertainty within the specified budget $\varepsilon_m$ by summing the failure probabilities $(1 - \hat{p}_{\sigma})$ of all assigned subtasks. Sequence consistency is maintained by constraint (Eq. S24), which ensures that if a robot is assigned a subtask ($x_{r,\sigma} = 1$), it must have exactly one subsequent subtask in its execution sequence. The non-overlapping constraint (Eq. S25) prevents cyclic dependencies in task sequencing by ensuring that two subtasks cannot be mutually consecutive in a robot's execution sequence. Causality is guaranteed by the start time constraint (Eq. S26), which ensures all subtasks start no earlier than the current decision time $t$. Finally, the variable domain constraints (Eq. S27) define the mathematical properties of decision variables, with binary assignment and sequencing variables, and non-negative continuous start time variables. This comprehensive formulation enables real-time subtask dispatching while addressing the practical constraints of multi-robot systems operating under the uncertainty. The formulated optimization problem is solved using off-the-shelf solver Gurobi (*60*), which efficiently handles the mixed-integer linear programming formulation. As described in the main text, our rolling horizon approach dispatches a limited number of subtasks at each decision epoch, maintaining a tractable problem size that enables the real-time computation. The overall procedure operates as a closed-loop system, i.e., a subtask graph is generated from high-level mission specifications, optimal dispatch decisions are computed under uncertainty constraints, and execution outcomes continuously update the task graph for subsequent cycles. This framework ensures continual adaptation to partial observability and dynamic environmental conditions while maintaining computational feasibility for real-world deployment.

## Human-in-the-Loop Verification, Monitoring and Online Adaptation

### *Automaton-based Monitoring with Correctness Guarantee*

As the main text illustrates, this component concerns the monitoring and verification of the collective execution. Mission automata are used to formally track the progress of each mission and guarantee the satisfaction of temporal specifications, while the induced partial ordering capture the synchronization constraints between tasks.

*(I) Monitoring via Trace Tracking.* Recall that each mission specification $\varphi_k$ is associated with a NBA $\mathcal{B}_{\varphi_k} = \left(Q^k, Q_0^k, \Sigma^k, \delta^k, Q_F^k\right)$. During execution, the trace $\rho_k\left(t\right) \in (\Sigma^k)^*$ is generated by the sequence of completed tasks. Monitoring requires updating the set of reachable states:

$$\hat{Q}_k\left(t+1\right) \triangleq \{q' \in Q^k \mid \exists q \in \hat{Q}_k\left(t\right), \sigma \in \Sigma^k : \left(q, \sigma, q'\right) \in \delta^k\} \tag{S28}$$

where $\sigma$ encodes the latest completed task. The mission $\varphi_k$ is satisfied if the trace visits final states infinitely often,

$$\rho_k \vDash \varphi_k \Leftrightarrow \lim_{t\to\infty} \sup \rho_k\left(t\right) \cap Q_F^k \neq \varnothing \tag{S29}$$

which provides a provable guarantee that the execution respects the LTL specification.

*(II) Synchronization via Dynamic Posets.* Based on $\mathcal{B}_{\varphi_k}$, the associated tasks and their temporal constraints can be abstracted into posets,

$$\Omega_\varphi \triangleq \left\{\left(\Omega, \leq, \simeq\right)\right\} \tag{S30}$$

where $\Omega = \left\{\omega_1, \cdots, \omega_M\right\}$ is the set of tasks, $\omega_{m_1} \leq \omega_{m_2}$ requires that $\omega_{m_1}$ finish before $\omega_{m_2}$ starts, and $\omega_{m_1} \simeq \omega_{m_2}$ enforces simultaneous start. During execution, these constraints translate to synchronization conditions such as

$$\max_{i\in\mathcal{C}_{\omega_{m_2}}} s_{m_2}^i \geq \min_{j\in\mathcal{C}_{\omega_{m_1}}} e_{m_1}^j \tag{S31}$$

for precedence, and

$$s_{m_1}^i = s_{m_2}^j, \forall i \in \mathcal{C}_{\omega_{m_1}}, j \in \mathcal{C}_{\omega_{m_2}} \tag{S32}$$

for simultaneity. Importantly, the poset $\Omega_\varphi$ evolves online as execution unfolds and new missions are added, so the set of enforced synchronization constraints is updated dynamically. Together, monitoring with mission automata (S28)-(S29) and synchronization with posets (S30)-(S32) yield a correctness guarantee: every feasible execution trace remains consistent with the mission specification, both in terms of temporal satisfaction and collaborative ordering.

## *Multi-level Online Adaptation*

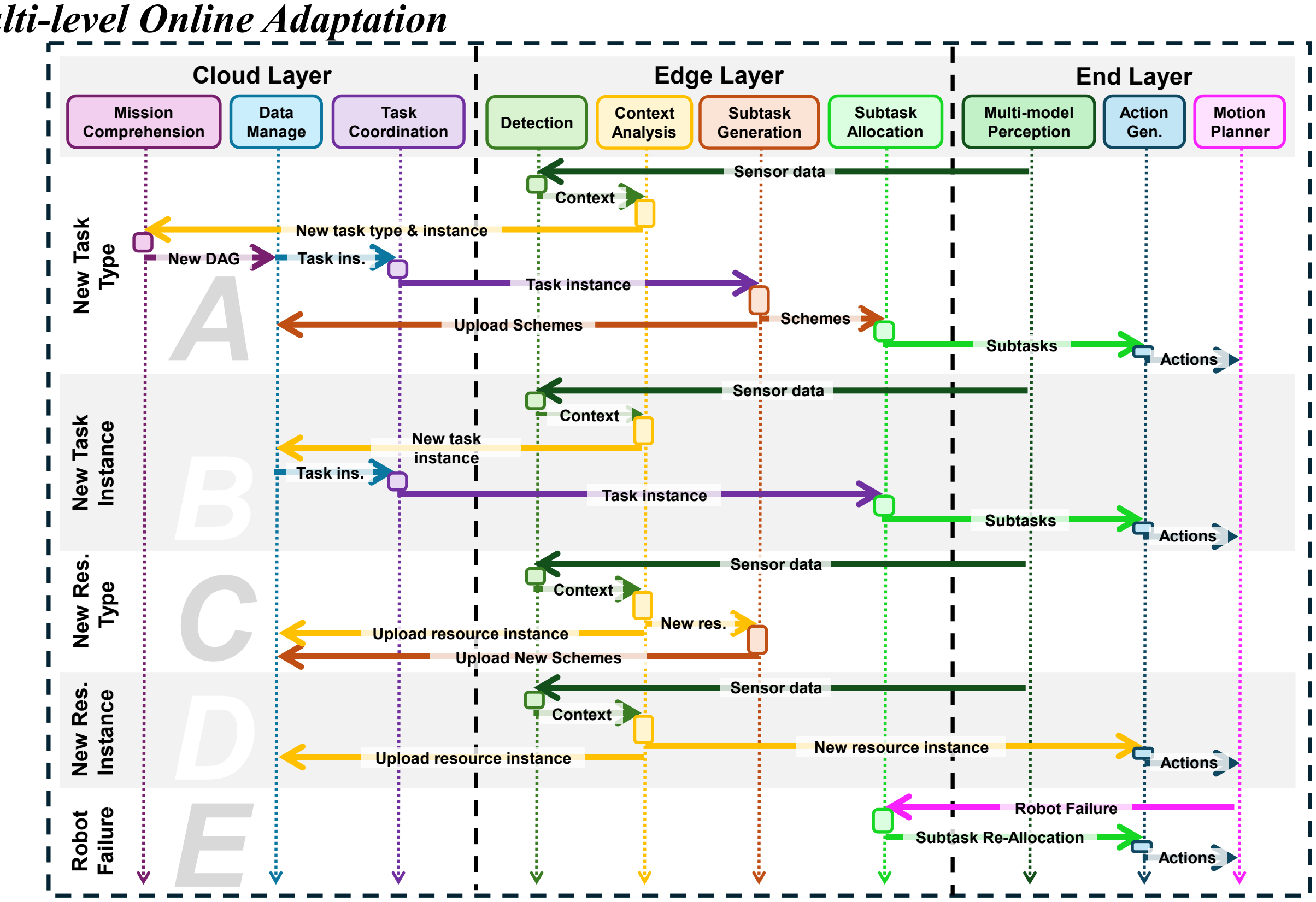

**Figure S7: Online adaptation mechanism in unknown environments.** System responses across five distinct scenarios: emergence of new task types, new task instances, new resource types, new resource instances, and robot failures. The coordinated reactions of system modules demonstrate robust adaptation capabilities in dynamic settings.

The ability to adapt to different contingencies online is essential for the swarm to operate autonomously in open and unknown environments. As illustrated in Fig. S7, our framework incorporates a multi-level online adaptation mechanism that leverages the cloud-edge-device architecture (detailed in the section of "Software Architecture" under Supplementary Results). This mechanism enables dynamic responses to the following representative events during the mission execution: (I) **New Task Type**. When new types of tasks are introduced in a mission, the cloud-layer task reasoning module is triggered to generate new LTL specifications and corresponding automata. The cloud then assigns these tasks to appropriate robot groups. Subsequently, the edge layer invokes a LLM to decompose the task into a sequence of subtasks, which are transmitted to the cloud database for updates. Finally, the subtask assignment module reallocates local plans for each robot. (II) **New Task Instances**. When new instances of known tasks are identified (e.g., newly detected survivors), the task allocation module at the cloud layer assigns the task to a specific group. The edge-layer subtask assignment module then redistributes the corresponding subtask sequence among the robots within the group. (III) **New Resource Type**. The detection of novel environmental features (e.g., the presence of water alongside a fire-extinguishing task) suggests potential new strategies for task decomposition. Consequently, the task reasoning module is re-triggered to update the layered DAG, which is forwarded to the subtask assignment module to revise local plans accordingly. The updated task schemes are also transmitted to the cloud database for the synchronization. (IV) **New Resource Instances**. When new instances of known features are observed (e.g., a new water reservoir is found), certain subtasks may be accomplished earlier. The system updates the resource information and uploads it to the cloud-layer data management module, making it accessible to other groups. (V) **Robot Failure**. In the event of a robot failure, the current plan becomes infeasible under unchanged global constraints. The subtask assignment module is reactivated to compute new feasible local plans for the remaining robots. This multi-level adaptation protocol ensures robust and responsive coordination across the swarm, enabling the continuous operation despite dynamic uncertainties in the environment.

## Hardware Experiment

### *Software Architecture*

Our software framework employs a cloud-edge-end architecture designed to modularize the system components across different layers, enabling both functional isolation and collaborative operation. As illustrated in Fig. S8, the cloud layer hosts two core modules: the mission comprehension module, which generates LTL specifications and the corresponding automata and task DAGs for newly identified missions, storing them in a cloud database for the shared access across all robot groups; and the task allocation module, which assigns new tasks to specific groups through automaton-based reasoning and Pareto-optimal tree search.

At the edge layer, each group operates a shared edge node—implemented on an operator-held tablet—that integrates four key modules: the *detection* module, which performs the semantic recognition of perceptual features collected from group robots; the *context analysis* module, responsible for interpreting semantic information as either tasks or resources; the *task reasoning* module, which reasons about newly discovered tasks and synthesizes executable subtask sequences; and the *subtask allocation* module, which distributes subtasks among group robots and generates local action plans. Notably, all modules at both cloud and edge layers support human verification and human-swarm interaction.

The end layer consists of individual robots, each equipped with a *perception* module (e.g., LiDAR or cameras). Upon receiving allocated subtask sequences, each robot selects appropriate action libraries for individual subtasks, derives concrete action sequences, and interfaces with a *motion planner* for physical execution. This hierarchical design efficiently supports large-scale robotic deployments: cloud components handle low-frequency updates and multi-group resource sharing, while edge intelligence enables the localized task processing within each group, minimizing inter-group communication overhead and enhancing intra-group coordination. When a group encounters tasks beyond its capability, the system escalates requests to the cloud for cross-group assistance. Consequently, edge modules are invoked frequently, whereas end-side intelligence operates in real time to execute subtasks independently. This layered invocation strategy ensures robust performance in highly dynamic and unknown environments, simplifying system complexity while maintaining operational coherence, particularly advantageous in large-scale applications.

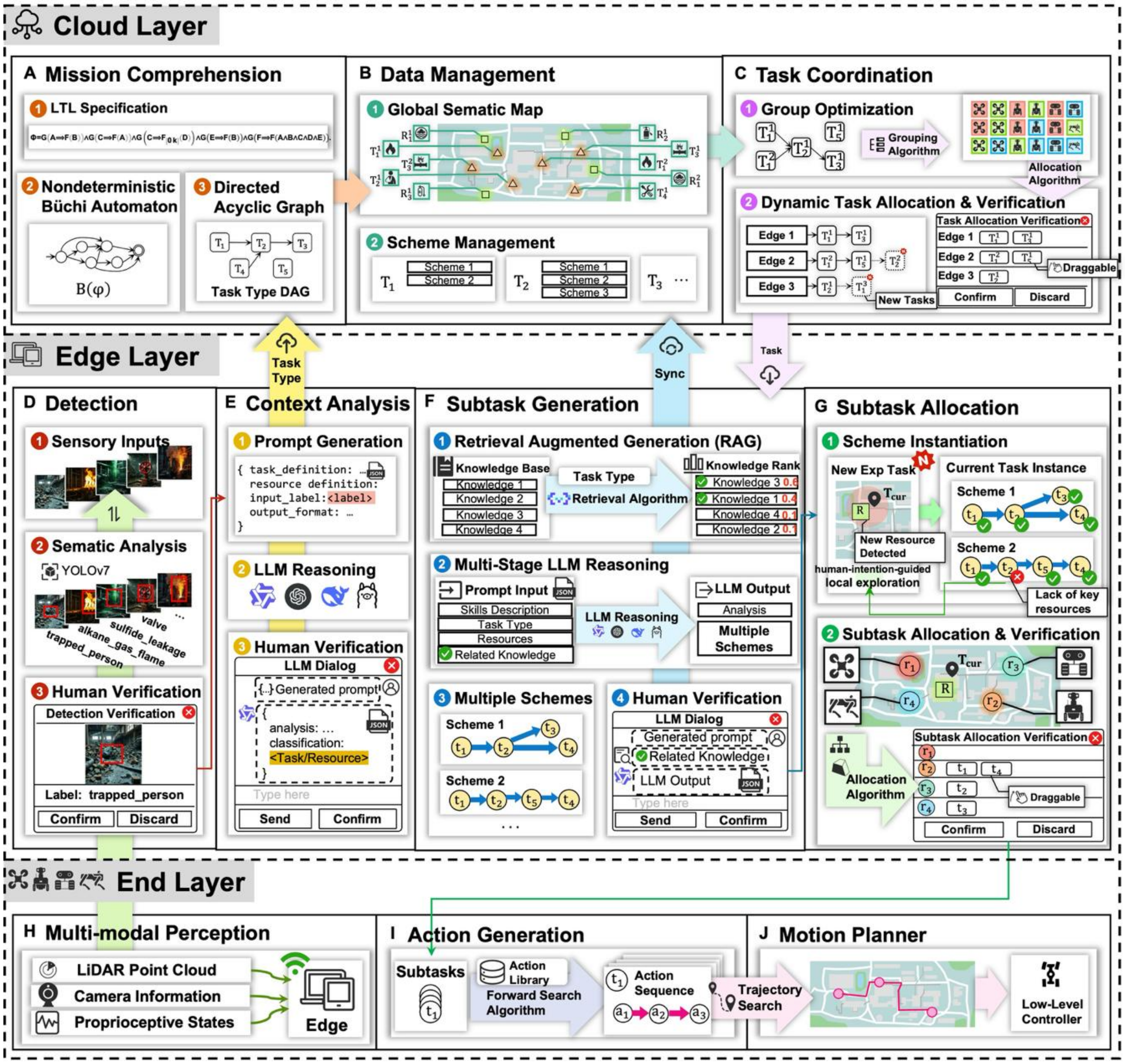

**Figure S8: System Architecture.** The cloud layer handles complex temporal planning using formal methods, maintains a global repository of tasks, resources, and schemes across all groups, and performs the task allocation via automata and Pareto-tree search. The edge layer decomposes tasks and allocates subtasks within each group through semantic perception, task/resource recognition, task reasoning, and scheduling modules. The end layer executes subtask sequences through local perception and motion planning, with robots following predefined action sequences while streaming real-time sensory data back to the edge.

## *Communication Network*

In large-scale heterogeneous multi-robot systems, communication architecture plays a critical role in system performance, as it determines connectivity patterns and communication overhead. Inefficient communication can impair information exchange within and between groups, ultimately compromising the cooperative task execution. To address this challenge, we design a specialized network architecture aligned with our cloud-edge-client framework. As shown in Fig. S9, the cloud components are deployed on a workstation equipped with 64GB RAM, 8-core CPU, and NVIDIA RTX 4090 GPU connected to a router-configured local area network (WI-FI 6) via ROS1. Due to computational constraints of handheld tablets, edge programs are also hosted on the same workstation, leveraging the internal ROS1 communication for efficient cloud-edge collaboration. To maintain operational control through tablets, we employ the Remote

Desktop Protocol (RDP), enabling each tablet to remotely operate edge programs while computational loads remain centralized at the workstation. This approach maintains the flexibility to deploy edge programs directly on tablets when necessary. For robot-edge communication, we implement specialized bridges to accommodate different robot platforms. Our unmanned aerial Vehicles (UAVs) and unmanned ground vehicles (UGVs) utilize ROS1 architecture, while the quadruped robots employ ROS2. To ensure interoperability, we adapt the Swarm ROS bridge package to encapsulate ROS messages into TCP/UDP formats for aerial and ground vehicles. For quadruped robots, we develop a TCP-based interface that encodes ROS1 messages into strings for transmission, which are subsequently decoded into ROS2 messages onboard the robots. This network design ensures reliable data exchange while maintaining platform-specific requirements. Each robot utilizes its internal network stack for local communication and motion execution. The implemented architecture demonstrates efficient cloud-edge-client coordination while supporting seamless collaboration across heterogeneous robot teams, thereby enabling robust task execution in complex environments.

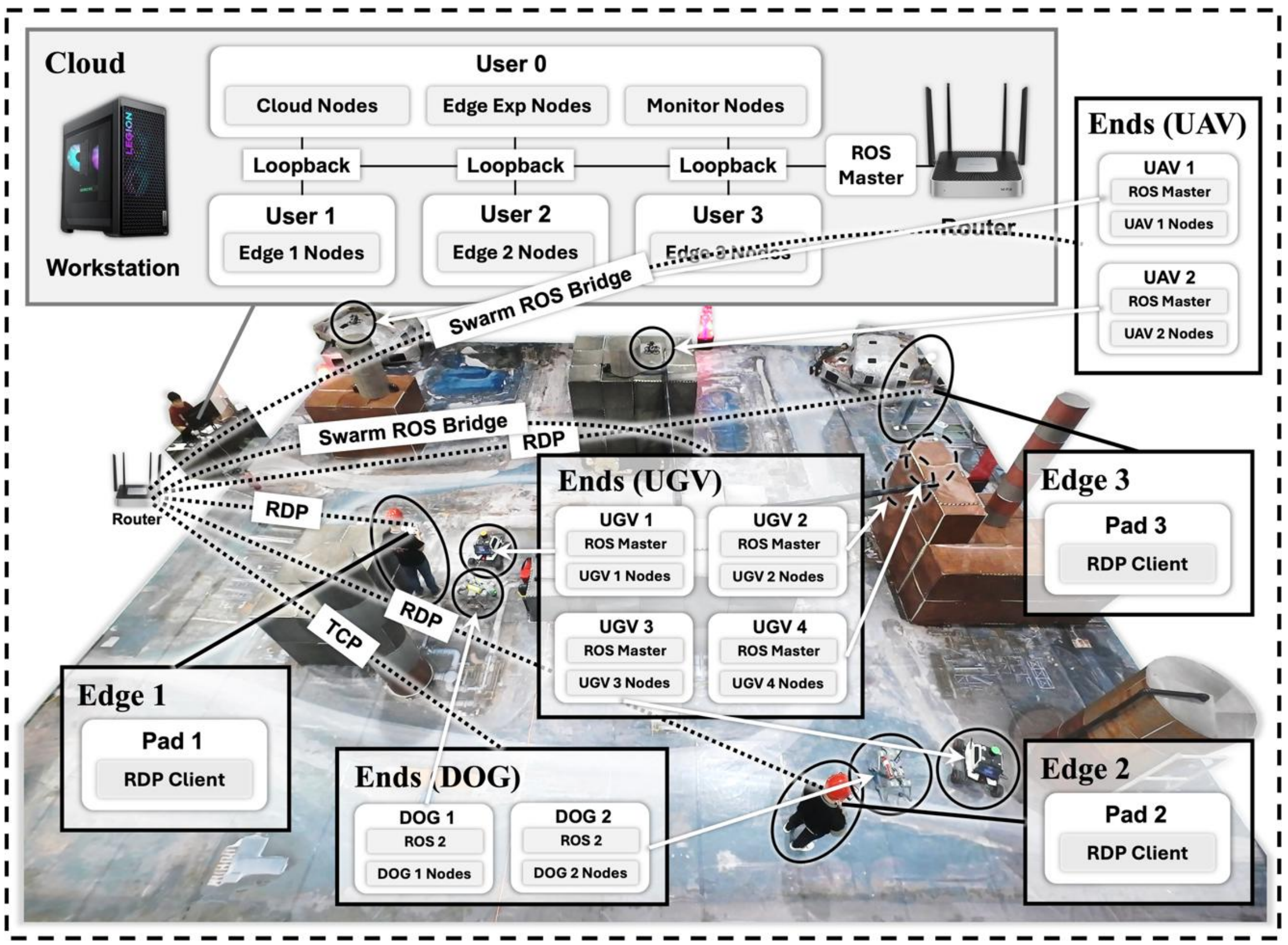

**Figure S9: System Architecture.** Communication architecture for cloud-edge-client robot coordination. Cloud and edge modules are co-hosted on a central workstation, with remote tablet control via RDP. Dedicated bridges enable the interoperability between ROS1-based UAVs/UGVs and ROS2-based quadrupeds, supporting the reliable data exchange across heterogeneous platforms.

## *Perception, Localization and Mapping*

Our heterogeneous robotic system employs customized perception and localization strategies tailored to the distinct capabilities of each platform—UAVs, UGVs, and quadruped robots. In terms of the perception, all robots utilize onboard stereo depth cameras to capture multi-view images. A unified semantic perception model is developed using YOLOv7 (*61*), trained on manually annotated datasets collected from all three platforms. Real-time inference is enabled via a ROS-compatible implementation of YOLOv7, ensuring the consistent semantic recognition across diverse viewpoints. Given the challenging indoor environment where positioning errors could lead to collisions, we establish a unified motion capture system with retro-reflective markers attached to each robot, forming trackable rigid bodies. Communication with the motion capture system is optimized using a TP-LINK-XVR3000L router operating at 120 Hz, enabling the high-precision pose estimation. This infrastructure ensures safe and stable robot operation in confined spaces, though our system remains extensible to outdoor environments with alternative localization sources such as GPS (*62*), RTK (*63*), or UWB (*64*).

For mapping and planning, platform-specific solutions are implemented. The UGVs employ FAST-LIO (*65*), a computationally efficient inertial-LiDAR odometry package, running on an NVIDIA Xavier onboard computer. A Livox MID-360 LiDAR provides 360° horizontal point cloud data at 20 Hz, fused with IMU measurements via an extended Kalman filter to generate accurate 3D maps. The quadruped robot utilizes the built-in 3D LiDAR with the slam-toolbox package (*66*), implementing the Gmapping package for real-time 2D map construction. Meanwhile, the UAVs leverage MID-360 LiDAR sensors and odometry sensors to build 3D Euclidean Signed Distance Field maps (*67*) for motion planning in three-dimensional space. This integrated approach ensures robust environmental perception, centimeter-level localization accuracy, and real-time map generation across all robotic platforms.

### *Robot Behaviors and Motion Planning*

Given the heterogeneity of our robotic platforms, customized motion planners are developed to ensure safe and reliable navigation to designated task areas. For UAVs, we implement the Ego Planner navigation package (*68*) for collision-free flight in three-dimensional environments, complemented by the coverage path planner for the global exploration and patrol missions (*69*). The UGVs utilize the ROS Move Base package for navigation (*70*), integrated with frontier-based exploration algorithms (*71*) to enable autonomous environmental exploration. The quadruped robot, operating exclusively on ROS2, employs the Navigation2 stack for collision-free navigation (*72*) with full pose control (*73*) including the target position, heading, and orientation. Additional specialized maneuvers such as stationary poses and circular trajectories are executed through dedicated API calls provided by the manufacturer. By integrating these specialized motion planners with our high-precision localization system, all robotic platforms can safely and reliably reach designated task areas while maintaining the system-wide operational safety and robustness.

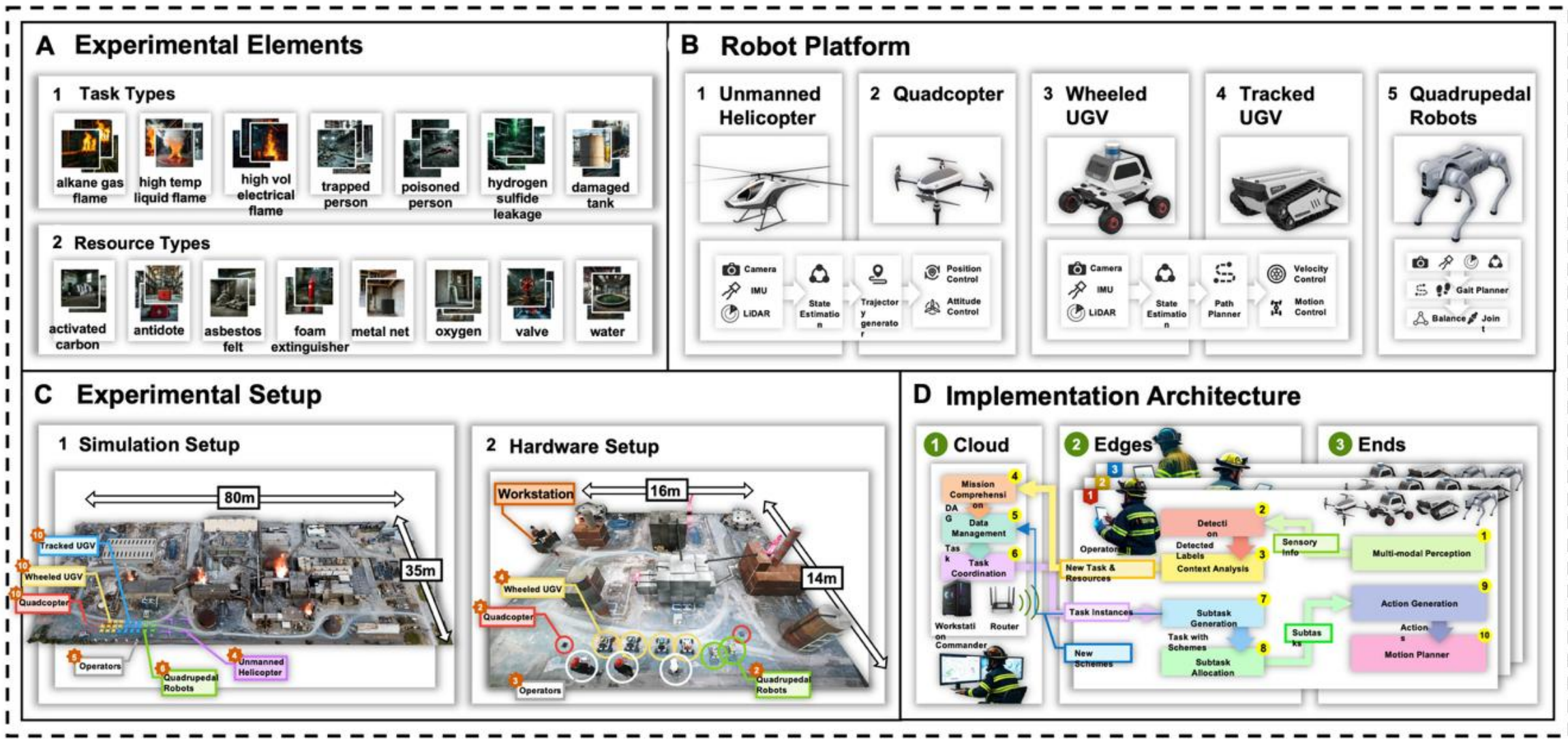

**Figure S10: Platforms and implementation setup. (A)** Visualization of task types and resource types. **(B)** Robot platforms and operational framework. **(C)** Physical experimental scenario and task distribution. **(D)** Cloud–edge–end deployment architecture.

**Table S2**: Robotic Capabilities and Specifications

| Type | Description | Velocity (m/s) | Skill Set |
|---|---|---|---|
| **UHeli** | Unmanned Helicopter | 2.0 | global_exploration |
| **UAV** | Unmanned Aerial Vehicle | 2.0 | local_exploration, inspection, monitoring, detection, throwing, liquid_spray, gas_spray, ignition |
| **UGV** | Unmanned Ground Vehicle | 2.0 | local_exploration, inspection, monitoring, transportation, ignition, solid_spray, liquid_spray, gas_spray |
| **TUGV** | Tracked UGV | 2.0 | local_exploration, inspection, monitoring, transportation, ignition, building, solid_spray, laying, cleanup, liquid_spray, gas_spray |
| **Dog** | Robotic Dog | 2.0 | local_exploration, inspection, monitoring, operation, building, rescue, cleanup, ignition, fixing |

## Supplementary Results

### *Additional Implementation Details*

To rigorously validate the effectiveness and performance of our proposed framework, comprehensive evaluations are conducted in both a large-scale high-fidelity simulation environment and physical deployments. The simulation environment comprises a 1000× 1000 m chemical plant facility while the physical scenario involves a real-world disaster response setting with size of 16×18 m, illustrated in Fig. S10 C, where four categories of emergency tasks dynamically emerge over a 11-minute operational period. These include fire suppression scenarios involving gas, liquid, and solid flames; casualty rescue operations with mild, severe, and critical injuries; chemical containment missions for gas, liquid, and solid leaks; and equipment protection tasks addressing mechanical failures and seepages. Table S3 enumerates each feature type, the basic skills required for completion, and the number of robots needed per skill. Entries marked as “N/A” indicate resource-type features that do not require active execution but must be carried by robots with corresponding skills to accomplish tasks. With these configurations, we assign suitable robots to each task, ensuring they possess the necessary skills and resources. This formulation naturally leads to a combinatorial optimization problem for subtask allocation, presenting a key challenge in our system implementation. To evaluate the framework’s capability in handling complex multi-robot coordination, each task requires collaborative execution by heterogeneous robots with complementary capabilities, as summarized in Table S2, that specifies the abbreviations, classifications, velocities, and skill sets for each robot type, highlighting their heterogeneous nature and capacity to perform distinct tasks.

Our experimental platform incorporates twenty heterogeneous robots spanning five specialized types: unmanned helicopters, UGVs, tracked UGVs, quadrupedal robots, and UAVs. As detailed in Fig. 2A, the swarm composition consists of three reconnaissance helicopters, six wheeled UGVs, four tracked UGVs, four quadrupeds, and four multirotors. While all platforms share fundamental capabilities including local_exploration, inspection, and monitoring, significant capability differentiation exists across robot types. Quadrupedal robots lack liquid and gas spraying functions but uniquely possess manipulation, triage, and repair skills. Tracked UGVs cannot perform ignition operations but exclusively support object placement functions. Multirotor UAVs and quadrupeds cannot handle solid spraying, while wheeled UGVs and multirotors lack construction and cleanup capabilities respectively. Successful task execution requires coordinated application of specialized resources including valves, switches, water reservoirs, foam suppressants, activated carbon, asbestos blankets, oxygen supplies, metal nets, and antidotes.

**Table S3:** Feature Types and Required Skills

| Feature Type | Basic Required Skills | Robot Count for Each Skill |
|---|---|---|
| **alkane_gas_flame** | inspect, operate, liquid_spray, monitor | 2, 1, 1, 2 |
| **high_temp_liquid_flame** | inspect, lay, liquid_spray, monitor | 2, 2, 1, 2 |
| **high-voltage_electrical_flame** | inspect, liquid_spray, monitor, operate, lay | 2, 1, 2, 1, 1 |
| **trapped_person** | inspect, clean_up, rescue, monitor | 2, 2, 1, 2 |
| **poisoned_person** | inspect, rescue, gas_spray, monitor | 2, 1, 1, 2 |
| **hydrogen_sulfide_leakage** | inspect, ignite, monitor, solid_spray | 2, 1, 2, 1 |
| **damaged_tank** | liquid_spray, monitor, fix | 1, 2, 1 |
| **valve** | N/A | N/A |
| **switch** | N/A | N/A |
| **water** | N/A | N/A |
| **foam** | N/A | N/A |
| **activated_carbon** | N/A | N/A |
| **asbestos_felt** | N/A | N/A |
| **oxygen** | N/A | N/A |
| **metal_net** | N/A | N/A |
| **antidote** | N/A | N/A |

All robot platforms implement fully autonomous navigation stacks for dynamic and unknown environments. Our physical validations employ Unitree Go2 quadrupeds (*74*), Scout unmanned ground vehicles (UGVs) (*75*), and custom unmanned aerial vehicles (UAVs). Crucially, all platforms implement identical algorithmic frameworks deployed directly on the robot hardware, as illustrated in Fig. S10 B, with the internal workflow for each robot detailed in the "Physical Experiment Deployment" section. To enable the efficient collaboration and distributed deployment, both our simulation and physical experiments adopt a unified cloud-edge-end architecture, depicted in Fig. S10D. This software architecture is described in detail in the "Software Architecture" section.

## *Additional Analysis on Planning Reliability*

In addition to the reliability metrics analyzed in the main text, we provide further measurements to evaluate the system robustness. First, we examine the impact of the adaptation mechanism defined as the capability to handle tasks with no initially available resources and to efficiently incorporate newly discovered ones. As shown in Fig. S11A, success rates with and without the exploration mechanism are compared across three difficulty levels. The results confirm that exploration significantly improves the task completion. Second, Fig. S11B demonstrates that our approach consistently reduces task completion time by flexibly utilizing alternative resources, highlighting the efficient adaptation to unexpected events. Furthermore, Fig. S11C presents the mean time between failures (MTBF) for all methods under varying difficulty scenarios. Our method achieves the longest MTBF, reflecting superior operational reliability. We also compare the length of generated plans against ground-truth plans of lengths 3, 4, and 5 across different difficulty levels. As illustrated in Fig. S11D, our method produces plans that closely match the ground truth in length, indicating appropriately scaled and human-interpretable solutions. Finally, Fig. S11E evaluates plan predictability, defined as the human ability to predict the full plan given a partial subtask sequence. Our method attains the highest predictability across all scenarios, underscoring its advantage in generating interpretable and predictable plans.

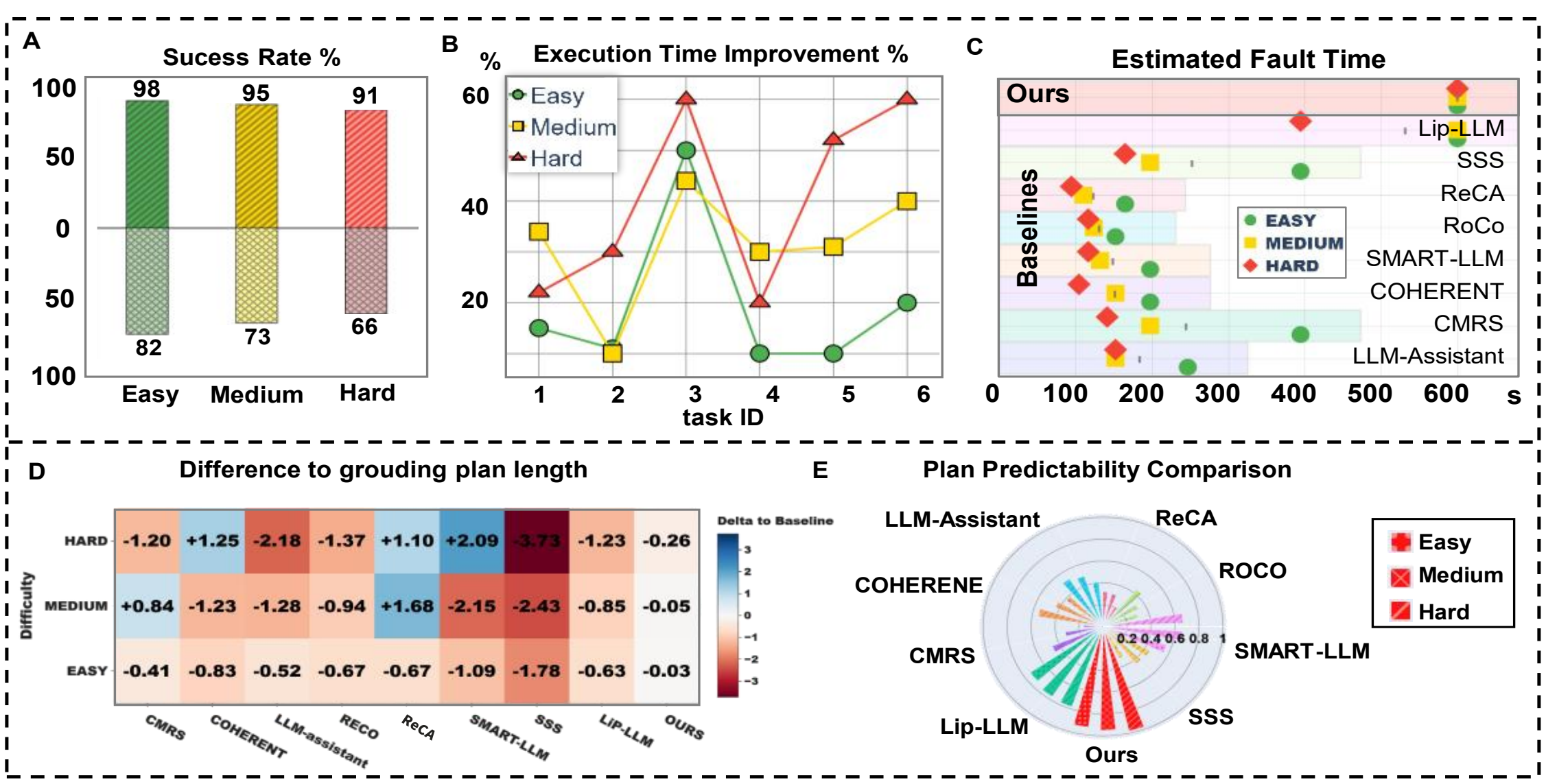

**Figure S11: Additional analysis on human-swarm interactions**. **(A)** Success rates with/without exploration across different difficulty scenarios. **(B)** Execution time reduction for tasks involving resource-type switching. **(C)** Mean time between failures under varying task difficulties. **(D)** Subtask sequence length deviation from ground truth is compared across difficulty levels, with smaller values indicating higher reliability. **(E)** Predictability of generated plans is assessed as the ability to infer remaining scheme segments, with higher scores corresponding to improve the reliability.

## *Additional Results on Human-swarm Interaction*

In the "Human-swarm Interaction" section of the main text, our system integrates six key human-swarm interaction modalities that collectively consider the human effort while preserving the valuable human oversight. These include: manual correction of detection results, editable modification of context analysis outputs, drag-and-drop adjustment of task allocations, predefined exploration areas, teleoperated robot skills, and voice-based interaction with LLMs. By incorporating these interaction channels, our system effectively embeds human intent into the autonomous pipeline, allowing domain expertise to refine algorithmically generated outcomes and optimize execution strategies. As demonstrated in Fig. S12, these interactions significantly enhance system performance. In Fig. S12 A, the system initially identifies an object as a fire extinguisher, but a human expert corrects it to a disinfectant through the detection interface. This correction enables the successful execution of a rescue mission for critically injured personnel, whereas the original misidentification would have prevented task completion. Fig. S12 B shows how the context analysis module mistakenly classifies green smoke as a resource, while expert intervention reclassifies it as toxic gas fire, enabling the appropriate reactive response. Task allocation flexibility is illustrated in Fig. S12 C, where the system initially assigns a quadruped robot to fetch water and a tracked vehicle for inspection. Through interactive reassignment, swapping these roles results in higher overall efficiency. For autonomous exploration, Fig. S12 D demonstrates how operators can define priority search areas based on situational awareness, guiding the tracked vehicle to focus on regions of expected value. The teleoperation capability shown in Fig. S12 E allows operators to monitor real-time camera feeds and directly execute robot skills. When the quadruped detects a leaking container, the operator triggers repair protocols through the interface, leveraging human judgment to preempt complex autonomous decision-making. Finally, Fig. S12 F highlights the integration of LLMs for the task reasoning, where text and voice inputs enable intuitive guidance of AI-generated solutions. These snapshots collectively validate the necessity of human-swarm interaction, demonstrating how combining human decision-making strengths with algorithmic autonomy yields more reliable, robust, and efficient system performance.

Additionally, we provide a comparative quantitative analysis beyond the human-swarm interaction metrics discussed in the main text. Figure S13A evaluates the contribution of each module in our system including: detection, task reasoning, and subtask scheduling, against fully manual planning in terms of the planning efficiency, time, and success rate. The results clearly demonstrate that each module significantly enhances planning efficiency across tasks of varying difficulty levels. Furthermore, Fig. S13B shows that with increased training time, both expert and non-expert users achieved substantially higher success rates and overall operational efficiency when using our system.

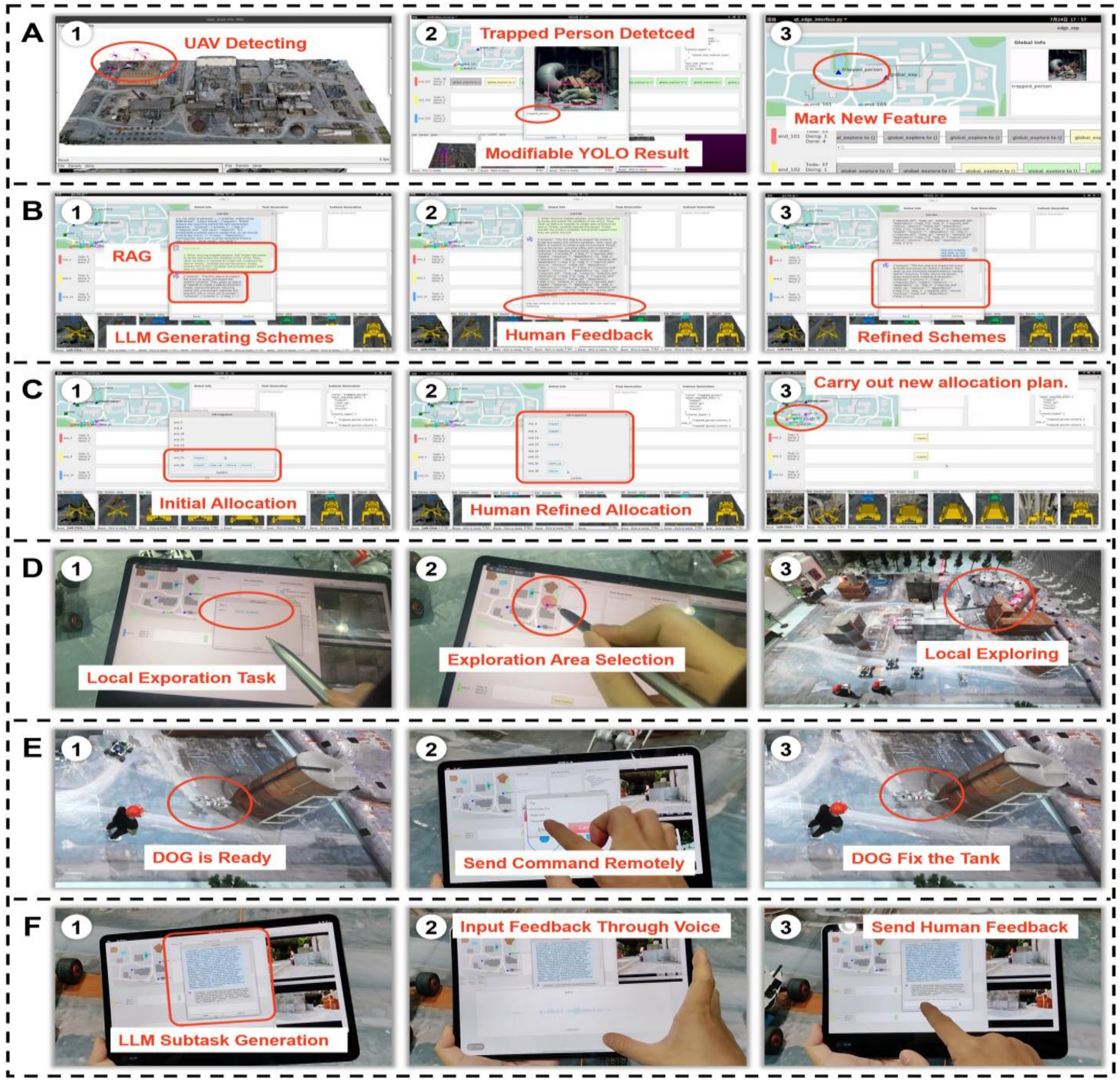

**Figure S12: Analysis on human-swarm interactions**. **(A)-(F)** Snapshots of six representative human-swarms interaction modalities, illustrating the corresponding before-and-after system states.

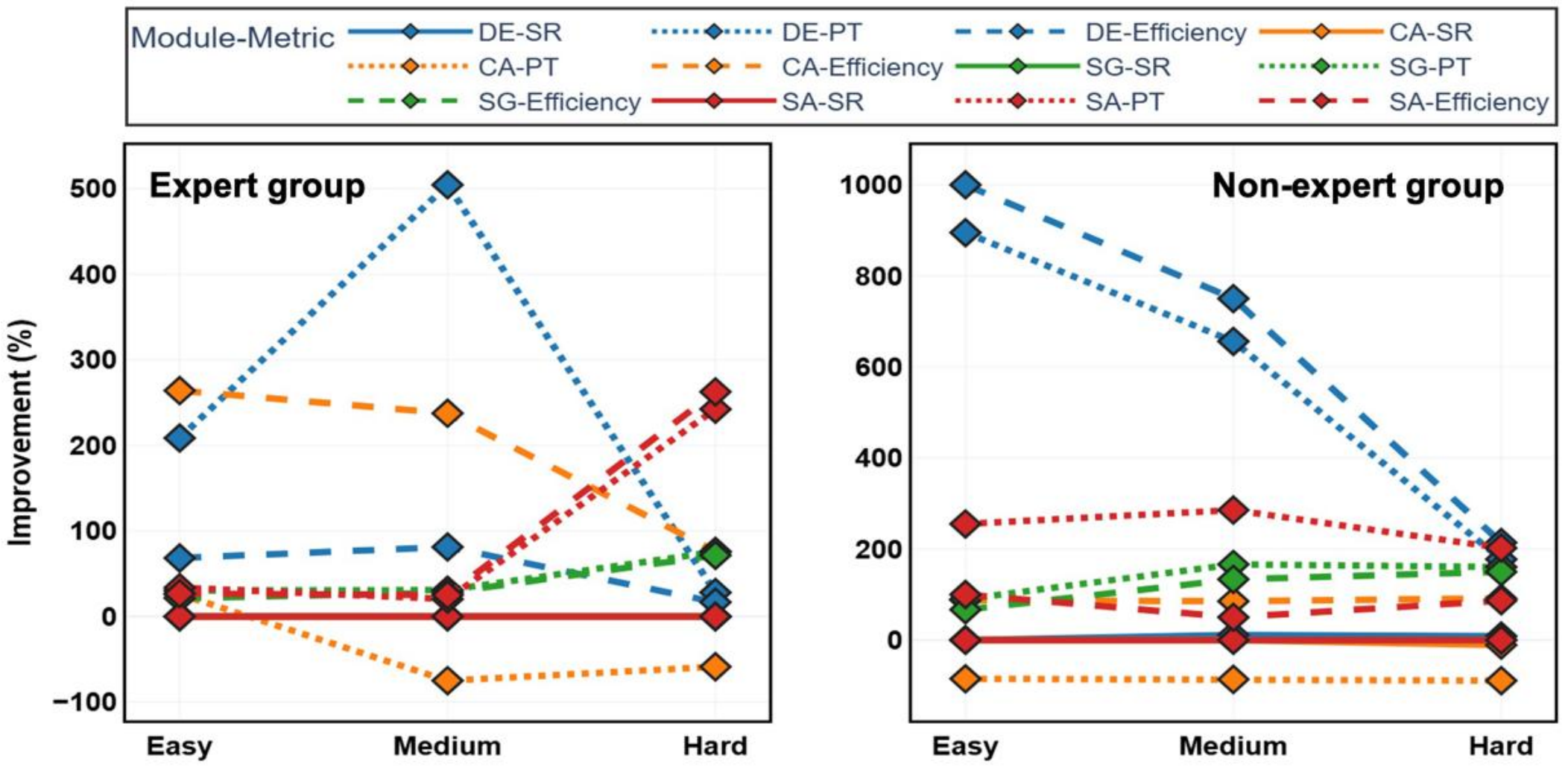

**Figure S13: Additional analysis on human-swarm interactions**. Improvement of each module in our system compared to fully manual planning.

## *Additional Results on Execution Monitoring*

As outlined in the "Materials and Methods" section, our system employs the task automata to monitor and verify the overall mission execution. The task automaton tracks the progress of each individual task, ensures compliance with temporal specifications, determines when each mission reaches completion, and predicts whether the ongoing execution sequence will lead to overall mission success. Fig. S14 illustrates a snapshot of currently active missions and their constituent tasks, displaying which tasks have been completed, which are currently executing, and which remain pending. The corresponding Gantt chart provides a complementary view of the same execution progress, showing the clear alignment with the automaton's state. This dual representation enables the intuitive and immediate assessment of mission status. For example, in Mission $\phi_1$ shown in Fig. S14, it is evident that three tasks remain before the full completion. All previously executed tasks have satisfied the LTL specifications, and the automaton confirms the existence of a feasible path to the accepting state, allowing early prediction that the mission will ultimately succeed. While the Gantt chart alone reveals task sequencing and timing, it cannot directly visualize whether the execution history adheres to complex temporal constraints, nor can it project future execution trajectories. The task automaton thus provides a formal and intuitive mechanism for the holistic mission supervision, offering both historical verification and prospective analysis of the entire operation.

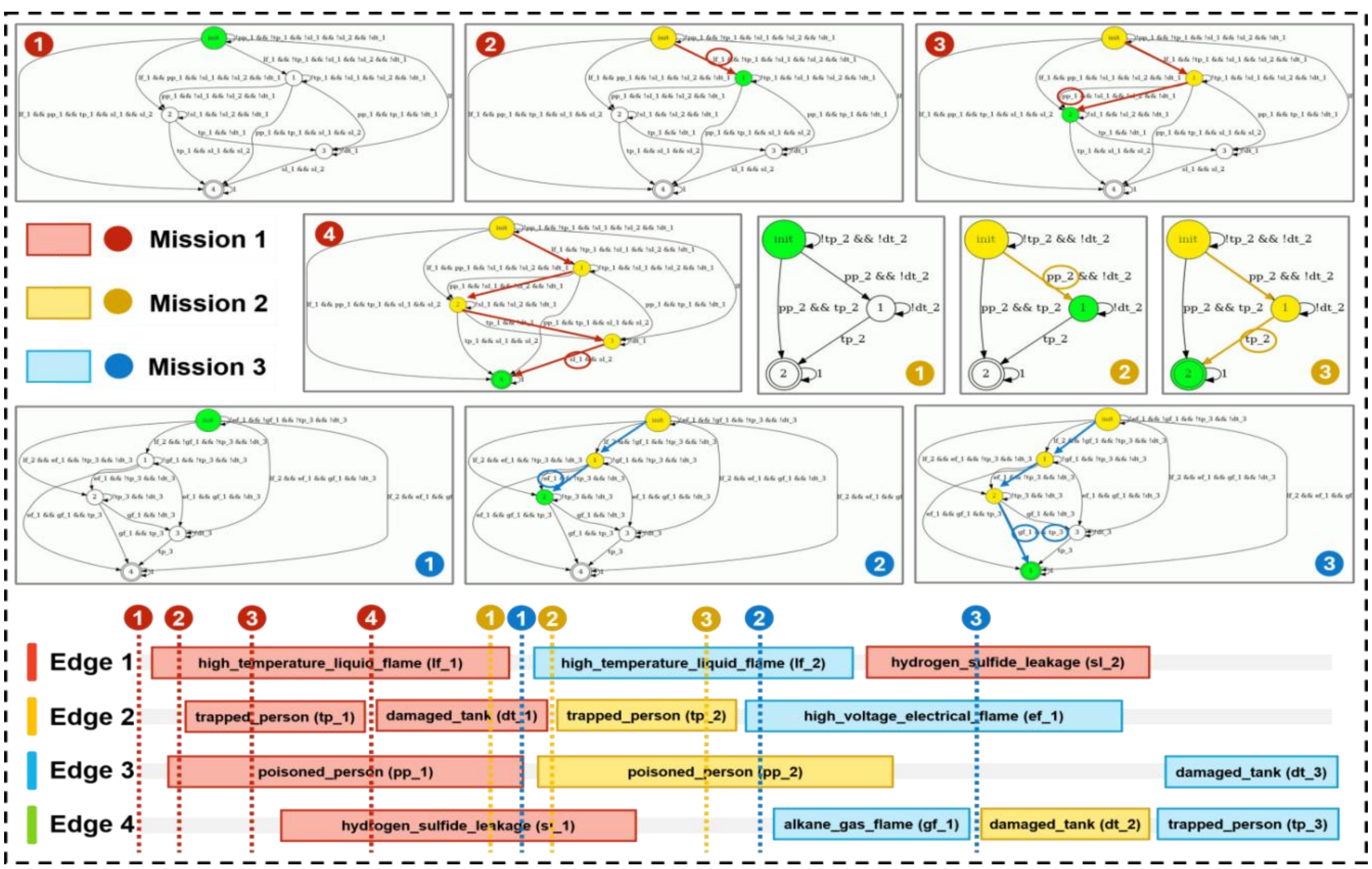

**Figure S14: Evolution of task automata during mission execution**. Automata evolution for three missions with the node color coding: green (executing tasks), gray (not started), and yellow (system state after completion). The red trace line shows the historical task execution. The Gantt chart below visualizes the group-level task progress, enabling the real-time monitoring of mission correctness and remaining states for successful completion.

## *Additional Results on Online Adaptation*

To enable adaptive and efficient responsiveness in unknown environments, our architecture incorporates a multi-level online adaptation mechanism within the cloud-edge-client framework, as detailed in the section of "Multi-level Online Adaptation" under Supplementary Methods. This approach allows different components of the system to react autonomously to various environmental changes, ensuring the continuous operational adaptation. Fig. S15 demonstrates our system's responsive capabilities across five representative scenarios encountered during experimental validation. In each case, the framework successfully triggers appropriate adaptive behaviors as designed, maintaining the robust and efficient performance under challenging conditions. Extensive tests confirm that these five core adaptation mechanisms suffice to handle the majority of real-world situations encountered during field operations, providing the comprehensive coverage for emergent environmental uncertainties while maintaining the system stability and mission continuity. Table S4 summarizes the average number of human-swarm interactions per task. The modules are not invoked uniformly; subtask scheduling is called most frequently, followed by subtask generation and then task reasoning. This pattern aligns with the designed priority of our online adaptation mechanism. Moreover, the increase in total interactions with task difficulty confirms the system's capacity for rapid real-time response.

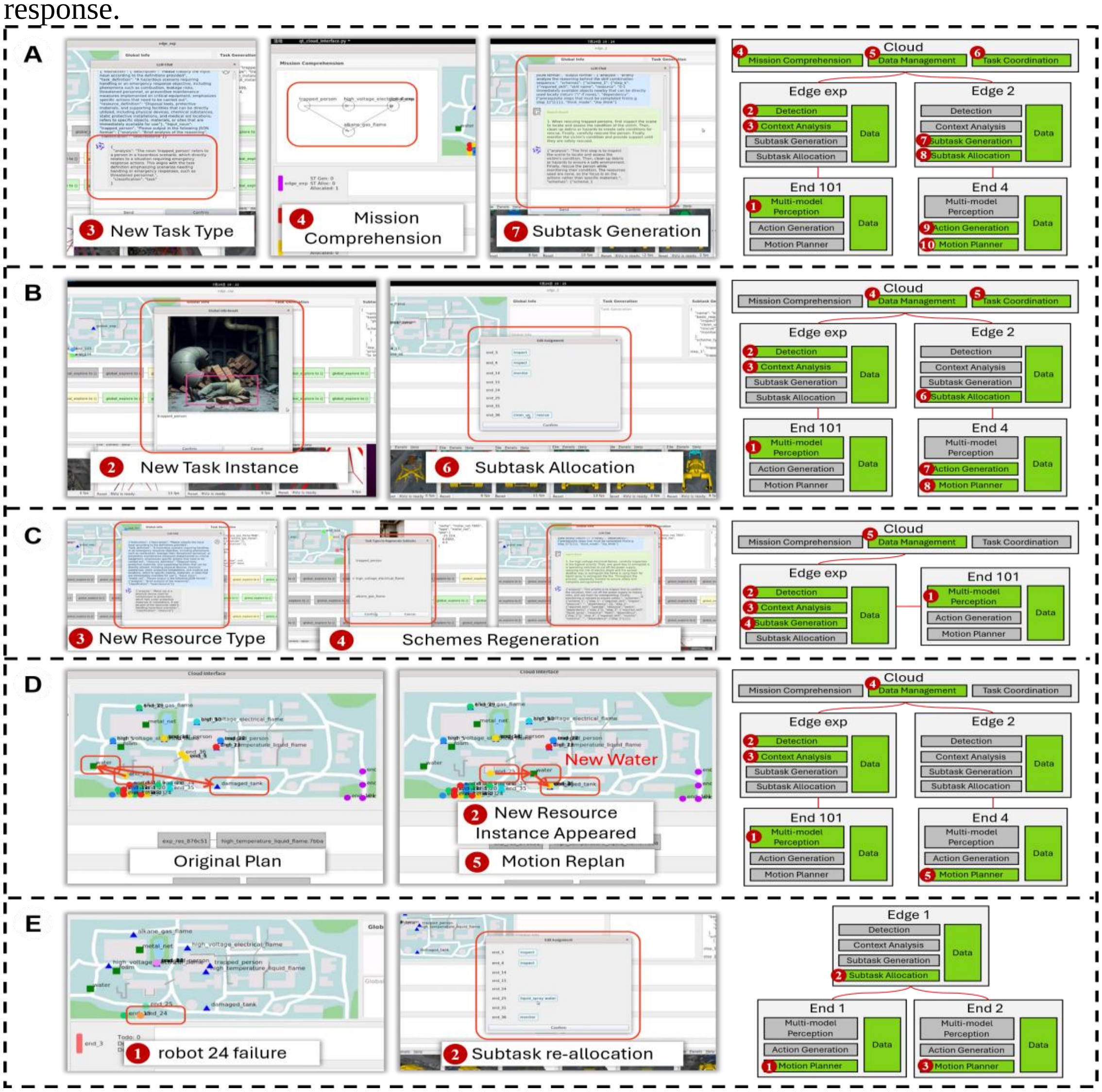

**Figure S15:** Snapshots of online adaptation scenarios. System responses across five distinct scenarios: emergence of new task types, new task instances, new resource types, new resource instances, and robot failures. The coordinated reactions of system modules demonstrate robust adaptation capabilities in dynamic settings.

**Table S4**: Number of module invocation under the online adaptation mechanism

| | Task Comprehension | Data Management | Task Coordination | Detection | Context Analysis | Task Reasoning | Subtask Scheduling |
|---|---|---|---|---|---|---|---|
| **Easy** | 1 | 42 | 21 | 34 | 15 | 8 | 40 |
| **Medium** | 11 | 53 | 20 | 45 | 18 | 15 | 49 |
| **Hard** | 15 | 66 | 20 | 60 | 27 | 25 | 60 |